\documentclass[journal]{IEEEtran}

\usepackage{graphicx}
\usepackage{setspace}
\usepackage{subcaption}
\usepackage{balance}
\usepackage{enumitem}
\usepackage{stfloats}
\usepackage{amsmath, amsthm, mathtools, amsfonts, amssymb}
\usepackage{ltxtable,tabularx, tabu,longtable, supertabular, makecell, multirow}
\usepackage[table]{xcolor}  
\usepackage{color}
\usepackage{pdfpages, pdflscape, everypage}
\usepackage{cite, url}
\usepackage{mathtools}
\usepackage{tabularx} 
\usepackage{array}
\usepackage{balance} 
\usepackage{pifont}

\usepackage{tikz}
\usetikzlibrary{trees}

\tikzstyle{box} = [rectangle, draw, minimum height=.9cm, rounded corners=.8ex, text centered]
\tikzstyle{big} = [text width=5.5cm]
\tikzstyle{grandchild} = [grow=down, xshift=-1cm, anchor=west, edge from parent path={(\tikzparentnode.200) |- (\tikzchildnode.west)}, text width=3.6cm]

\definecolor{yello}{RGB}{255,230,153}
\definecolor{orang}{RGB}{248,203,173}
\definecolor{grenn}{RGB}{197,224,180}
\definecolor{bluee}{RGB}{189,215,238}

\tikzstyle{first} = [level distance=1.2cm]
\tikzstyle{second} = [level distance=2.4cm]
\tikzstyle{third} = [level distance=3.6cm]
\tikzstyle{forth} = [level distance=4.8cm]
\tikzstyle{fifth} = [level distance=6cm]
\tikzstyle{level 1} = [sibling distance=5cm]

\definecolor{mycolor}{rgb}{1,0.0,0.5}

\newcommand{\Lpagenumber}{\ifdim\textwidth=\linewidth\else\bgroup
  \dimendef\margin=0 
  \ifodd\value{page}\margin=\oddsidemargin
  \else\margin=\evensidemargin
  \fi
  \raisebox{\dimexpr-\topmargin-\headheight-\headsep-0.5\linewidth}[0pt][0pt]{%
    \rlap{\hspace{\dimexpr \margin+\textheight+\footskip}%
    \llap{\rotatebox{90}{\thepage}}}}%
\egroup\fi}
\AddEverypageHook{\Lpagenumber}%
\begin{document}
\title{Games of GANs: Game-Theoretical Models for Generative Adversarial Networks}

\author{{ \normalsize Monireh Mohebbi Moghadam$^\dagger$, Bahar Boroomand$^\dagger$$^*$, Mohammad Jalali$^\dagger$$^{*}$, Arman Zareian$^\dagger$$^{*}$, \\ Alireza Daeijavad$^\dagger$, Mohammad Hossein Manshaei$^\ddagger$, and Marwan Krunz$^\ddagger$ \\ $^\dagger$Department of Electrical and Computer Engineering, Isfahan University of Technology, Isfahan, Iran \\
$^\ddagger$Department of Electrical and Computer Engineering, University of Arizona, Tucson, USA.
\thanks{$^{*}$These authors contributed equally.} }}

\maketitle
\begin{abstract}
Generative Adversarial Networks (GANs) have recently attracted considerable attention in the AI community due to its ability to generate high-quality data of significant statistical resemblence to real data. Fundamentally, GAN is a game between two neural networks trained in an adversarial manner to reach a zero-sum Nash equilibrium profile. Despite the improvement accomplished in GANs in the last few years, several issues remain to be solved. 
This paper reviews the literature on the game theoretic aspects of GANs and addresses how game theory models can address specific challenges of generative model and improve the GAN's performance. We first present some preliminaries, including the basic GAN model and some game theory background. We then present taxonomy to classify state-of-the-art solutions into three main categories: modified game models, modified architectures, and modified learning methods. The classification is based on modifications made to the basic GAN model by proposed game-theoretic approaches in the literature. We then explore the objectives of each category and discuss recent works in each category. Finally, we discuss the remaining challenges in this field and present future research directions. 
\end{abstract}
\begin{IEEEkeywords}
\label{KeyWords}
Generative Adversarial Network (GAN), Game Theory, Multi-Agent Systems, Deep Generative Models, Deep Learning.
\end{IEEEkeywords}

\section{Introduction}
\label{Sec:Intro}
Generative Adversarial Networks (GANs) represent a class of generative models that was originally proposed by Goodfellow et. al. in 2014 \cite{goodfellow2014GANarxiv}. They have received wide attention in recent years due to their potential to model high-dimensional complex real-world data~\cite{hong2019CSUR}.
As generative models, GANs do not minimize a single training criterion. They are used to estimate the real data probability distribution. A GAN usually comprises two neural networks: a discriminator and a generator, which are trained simultaneously via an adversarial learning technique. Such a GAN is more powerful in both feature learning and representation \cite{fedus2017arXiv}. The discriminator attempts to differentiate between real data samples and fake samples made by the generator, while the generator tries to create realistic samples that cannot be distinguished by the discriminator \cite{li2018Elsevier}, as shown in \figurename~\ref{Fig:simpleGAN}.

\tikzstyle{simple} = [rectangle, text width=2.2cm, minimum width=2cm, minimum height=.5cm,text centered, draw=black]

\tikzstyle{round} = [rectangle, rounded corners=12, text width=1.2cm, minimum height=1cm,text centered, draw=black]

\begin{figure*}[t]
	\centering	
	\begin{tikzpicture}[node distance=2cm]
		\node (noise) [simple, text width=1.8cm] {Noise $z$};
		\node (gene) [simple, right of=noise, xshift=1cm] {Generator $G$};
		\node (real) [simple, above of=gene, yshift=-1cm] {Real samples};
		\node (empty) [coordinate, right of=gene, xshift=.5cm, yshift=.5cm] {};
		\node (disc) [simple, right of=empty, text width=3cm, xshift=.5cm, minimum height=1.5cm] {Discriminator $D$};
		\node (emptydisc) [coordinate, below of=disc, yshift=0.5cm] {};
		\node (true) [round, right of=disc ,xshift=1cm] {True/ Flase};
		
		\draw [-latex] (noise) -- (gene);
		\draw (gene) -| node[anchor=335] {$G(z)$} (empty);
		\draw (real) -| node[anchor=330] {$x\quad$} (empty);
		\draw [-latex] (empty) -- (disc);
		\draw [-latex] (disc) -- (true);
		\draw [dotted] (true) |- (emptydisc);
		\draw [dotted,-latex] (emptydisc) -- (disc.270);
		\draw [dotted,-latex] (emptydisc) -| (gene);		
	\end{tikzpicture}
	\caption{GAN's basic architecture \cite{wang2017IEEE}.}
	\label{Fig:simpleGAN}
\end{figure*}
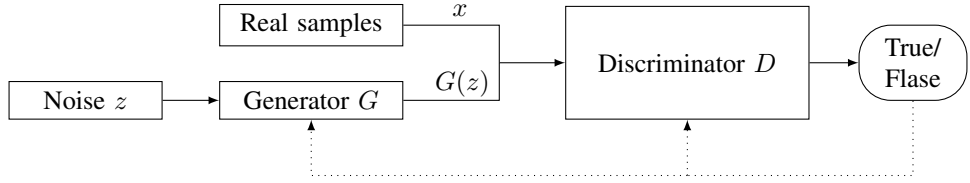	

In particular, many GAN models do not rely on any assumptions about the distribution of the real data they are trying to mimic. They can generate infinitely realistic new samples from latent space \cite{hong2019CSUR}. This feature enables GANs to be successfully applied in various applications, ranging from image synthesis, computer vision, video and animation generation, to speech and language processing, and cybersecurity \cite{alqahtani2019Springer}.

The core idea of GANs is inspired by a two-player zero-sum minimax game between the discriminator and the generator. In such a game, the total utilities of two players are zero, and each player’s gain or loss is exactly balanced by the loss or gain  of the other player. GANs are designed to reach a Nash equilibrium at which each player cannot increase their gain without reducing the other player's gain \cite{goodfellow2014GANarxiv, wang2017IEEE}.
%
Despite the significant success of GANs in many domains, applying them to certain real-world problems has been hindered by various challenges. The most significant of these problems is that GANs are that they are hard to train and suffer from instability problems, such as mode collapse, non-convergence, and vanishing gradients. A GAN needs to converge to the Nash equilibrium during the training process, but such convergence has been shown to be challenging~\cite{wang2019arXiv, wiatrak2019arXiv}.

Since their introduction in 2014, GANs have been widely studied, and numerous methods have been proposed to address their challenges.
However, to synthesize high-quality data using GANs, it is necessary to improve their underlying theory, which has been a major hindrance towards developing GANs~\cite{cao2018IEEE}. As the basic principle of GANs is based on game theory and the data distribution is learned via a game between the generator and discriminator, exploiting the game-theoretical techniques became one of the most discussed topics, attracted research efforts in recent years.

\subsection{Motivation and Contribution}

The primary motivation behind this survey is the absence of others reviews that focus on the game-theoretical advances in GANs. Other published surveys (e.g. \cite{goodfellow2016arXiv, wang2017IEEE, creswell2018IEEE, hitawala2018arXiv, gonog2019ICIEA, hong2019CSUR, bissoto2019arXiv, zhang2018IEEE, pan2019IEEE, kumar2020Springer, salehi2020arXiv, ghosh2020ICCCNT, saxena2020arXiv, gui2020arXiv, jabbar2020arXiv}) considered other aspects of GANs, as explain later. To the best of our knowledge, this work is the first to explore the GAN advancement from a game-theoretical perspective. 

 
Our survey first introduces some background and key concepts in this field. Then we classify recently proposed {\it games of GANs} models into three major categories: (i) modified game models, (ii) modified architectures in terms of the number of agents, and (iii) modified learning methods. We then classify each group into several subcategories. We review the main contributions of each work in each subcategory. We also point to some existing problems in the discussed literature and highlight possible future research directions. 

\subsection{Paper Structure and Organization}
The rest of this paper is organized as follows. Section \ref{Sec:Background} presents some background on game theory, and GANs includes basic idea, learning method, and challenges. In Section \ref{Sec:Related_Works} we have a glimpse to the other surveys conducted in the field of GANs. We provide our proposed taxonomy in Section \ref{Sec:Taxonomy} and review the research models in each category in this section. The final section is devoted to discussion and conclusions.


\section{Background and Preliminaries}
\label{Sec:Background}

\definecolor{tableShade}{gray}{0.88}

We start by an overview of game theory and then move toward GANs. Table \ref{Tab:Abbreviations} lists the acronyms and their definitions used throughout the paper.
\begin{table}[t]
    \rowcolors{2}{tableShade}{white}
	\centering
	\caption{Acronyms and corresponding full names appearing in the paper.}
	\begin{tabularx}{\columnwidth}{|l|X|}
		\hline
		Acronym& Full Name \\ \hline 
		GAN & Generative Adversarial Network\\ 
		WGAN & Wasserstein GAN\\
		MAD-GAN & Multi-Agent Diverse GAN \\
		MADGAN & Multiagent Distributed GAN \\
		MPM GAN& Message Passing Multi-Agent GAN \\
		Seq-GAN & Sequence GAN\\
		L-GAN& Latent-space GAN \\
		FedGAN & Federated GAN \\
		ORGAN & Objective-Reinforced GAN \\
		CS-GAN &  Cyclic-Synthesized GAN\\
		SCH-GAN & Semi-supervised Cross-modal Hashing GAN\\
		MolGAN & Molecular GAN\\
		RNN & Recurrent Neural Network\\
		RL & Reinforcement Learning \\
		IRL & Inverse Reinforcement Learning \\
		NE & Nash Equilibrium\\
		JSD& Jensen-Shannon Divergence \\
		KL & Kullback-Leibler \\
		AE& Auto-Encoder\\
		DDPG& Deep Deterministic Policy Gradient \\
		ODE & Ordinary Differential Equation \\
		OCR & Optical Character Recognition\\
		SS & Self-Supervised task \\
		MS & Multi-class minimax game based Self-supervised tasks \\
		SPE & Subgame Perfect Equilibrium\\
		SNEP & Stochastic Nash Equilibrium Problem\\
		SVI & Stochastic Variational Inequality\\
		SRFB & Stochastic Relaxed Forward-Backward\\
		aSRFB & Averaging over Decision Variables\\
		SGD & Stochastic Gradient Descent\\
		NAS & Neural Architecture Search \\
		IID & Independent and Identically Distributed \\
		DDL & Discriminator Discrepancy Loss\\
		
		\hline
	\end{tabularx}
	\label{Tab:Abbreviations}
\end{table}
\subsection{Game Theory} 
Game theory aims to model the interactions between several decision makers. These interactions called a "game" and the decision makers are called "players". In each round of the game, players take actions from a set of actions called strategy set. It is usually assumed that players are rational, which means that each one of them tries to maximize its own utility by selecting the action that maximizes its payoff. The action of a player is chosen with respect to other players' actions. because of that, each agent should have a belief about the other players \cite{osborne2004introduction}.

Several solution habe been introduced for analyzing games and finding their points of equilibrium. One type of equlibrium is "Nash equilibrium" at which each player cannot increase its payoff by changing its strategy unilaterally. In the other words, Nash equilibrium is a state where nobody regrets about its choice given others' strategies and with respect to its own payoff \cite{shoham2008multiagent}. In the situation where the players assign a probability distribution to the strategy sets instead of choosing one strategy, the Nash equilibrium is called \emph{mixed Nash equilibrium} \cite{shoham2008multiagent}. Constant-sum games are two player games in which sum of the two players' utilities is equal to this amount in all other states. When this amount equals to zero, the game is called \emph{zero-sum game} \cite{shoham2008multiagent}.

Considering zero-sum games, we can define another solution concept named maximin or minimax strategy. In the maximin strategy, the decision maker maximizes its worst-case payoff, which happens when all other players cause as much harm as they can to the decision maker. In minimax strategy, the decision maker wants to cause harm to others by minimizing other players' maximum payoff \cite{shoham2008multiagent}. The value which players get in the minimax or maximin strategy method is called min-max (minimax) or max-min (maximin) value, respectively. In \cite{neumann1928theorie}, Neumann proved that in any finite, two-player, zero-sum games, all Nash equilibria coincides with min-max and max-min strategies of players. Furthermore, the min-max and max-min values are equal to the Nash equilibrium utility. This is the most crucial result of zero-sum games that applies to generative adversarial networks.  
\subsection{Generative Adversarial Networks}
In this section, we give a brief description of the GAN by reviewing its basic idea, learning methods, and its challenges.

\subsubsection{Generative Models}

The main purpose of a generative model is to generate synthetic data whose distribution captures the distribution of a real data set (i.e., the training set). Generative models can be divided into three types. In the first type, the model is trained using a training set with distribution $p_{data}$ (unknown to model). The model generates data of distribution $p_{model}$, which is an approximation of $p_{data}$. In the second type, the model is solely capable of producing samples from $p_{model}$. The third type can do both approaches. GANs concentrate on sample generation, but it is also possible to design GANs that can do both approaches~\cite{goodfellow2016arXiv}.

\subsubsection{GAN Fundamentals}
In 2014, Goodfellow et al. \cite{goodfellow2014generative} introduced GANs as a framework in which two neural network (as players) play a zero-sum game. In this game, the players are called the generator $G$ and the discriminator $D$. The generator is the one that produces the samples while the discriminator tries to distinguish training samples from the generator's samples. The more indistinguishable the samples produced are, the better is the generative model~\cite{goodfellow2014generative}. Any differentiable function, such as a multi-layer neural network, can represent generator and discriminator. The generator, denoted by $G(z)$, takes as input a prior noise distribution ($p_{z}$) and maps it to approximate training data distribution ($p_{g}$). Discriminator, $D(x)$, is a mapping of the input data distribution $p_{data}$ into a real number in the interval of $[0,1]$, which represents the probability that the sample is real and not a fake one (i.e., produced by the generator)~\cite{gonog2019ICIEA}.
\subsubsection{GAN Learning Models}
The generator and discriminator can be trained using an iterative process for optimizing an objective function. Goodfellow et al. used the following objective function in their game~\cite{goodfellow2014generative}:

\begin{multline}
	\min\limits_{G} \max\limits_{D} V(G, D)
	=\mathbb{E}_{x \sim p_{data}(x)} [f_{0}(D(x))]\\
	+\mathbb{E}_{z \sim p_{z}(z)} [f_{1}(1-D(G(z)))]
	\label{eq:minimaxlog}
\end{multline}

where $ f_{0} $ and $ f_{1} $ take different forms, depending on the specific divergence metrics, as shown in \tablename~\ref{Tab:divergance}. The first proposed GAN uses the Jensen-Shannon divergence metric.
\setlength{\tabcolsep}{5pt}
\begin{table}
	\caption{Variant of GANs based on different divergence metrics\cite{ge2018ECCV}.}
	\label{Tab:divergance}
	\centering
	\begin{tabularx}{\columnwidth}{|c|c|c|c|}
		\hline
		{\bf Divergence metric}	& $f_{0}(D)$	& $ f_{1}(D) $		&  {\bf Game Value } \\
		\hline \hline
		Kullback-Leibler	& $ \log(D) $	& $ 1 - D $ 		&  $ 0 $ \\
		\hline
		Reverse KL			& $ -D $		& $ \log(D) $ 		& $ -1 $ \\
		\hline
		Jensen-Shannon 		& $ \log(D) $ 	& $ \log(1 - D) $	& $ -\log4 $ \\
		\hline
		WGAN 				& $ D $ 		& $ -D $			& $ 0 $ \\
		\hline
	\end{tabularx}
\end{table}
\setlength{\tabcolsep}{6pt}

To train the simple model, shown in \figurename~\ref{Fig:simpleGAN}, we first fix $G$ and optimize $D$ to make it discriminate as accurately as possible. Next, we fix $D$ and try to minimize the objective function. Discriminator operates optimally if it cannot distinguish between real and fake data. For example, under the Jensen-Shannon divergence metric, when $p_{r}(x)/(p_{r}(x) + p_{r}(z)) = 0.5$, if both discriminator and generator work optimally, the game reaches the Nash equilibrium and the min-max and max-min values  are the same (i.e., $-log4$ as shown in \tablename~\ref{Tab:divergance}). 


\section{Related Surveys}
\label{Sec:Related_Works}

As GANs became increasingly popular, numerous surveys have been presented (about 40 surveys so far), which can be classified into three categories. 
The literatures in the first category (\cite{goodfellow2016arXiv, wang2017IEEE, creswell2018IEEE, hitawala2018arXiv, gonog2019ICIEA, hong2019CSUR, bissoto2019arXiv, zhang2018IEEE, pan2019IEEE, kumar2020Springer, salehi2020arXiv, ghosh2020ICCCNT, saxena2020arXiv, gui2020arXiv, jabbar2020arXiv}) explore a relatively broad scope in GANs, including key concepts, algorithms, applications, different variants and architectures.
In contrast, the surveys in the second group ( \cite{lucic2017gans, alqahtani2019Springer, wiatrak2019arXiv, lee2020arXiv, pan2020IEEE}) focus on a specific issue in GANs (e.g., regularization methods, loss functions, etc) and address how researchers deal with such an issue. In the third category (\cite{cao2018IEEE, wang2019arXiv, wu2017TUP, wang2020IEEEAccess, sorin2020Elsevier, tschuchnig2020Patterns, agnese2020Wiley, jain2020JCR, sampath2020, yi2019Elsevier, yinka2019Springer, ghosh2020ICCCNT, di2019arXiv, geogres2020, gao2020arXiv, shin2020ICACT}) the surveys summarize the application of GAN in a specific field, from computer vision and image synthesis, to cybersecurity and anomaly detection. In the following, we briefly review surveys in each category and explain how our paper differs from them.
\subsection{GAN General Surveys}
In \cite{goodfellow2016arXiv}, The author provides answers to the most frequently raised questions about GANs. 
Wang et al. \cite{wang2017IEEE} reviewed theoretic and implementation models of GANs, their applications, as well as the advantages and disadvantages of this generative model. 
Creswell et al. \cite{creswell2018IEEE} provided an overview of GANs, especially for the signal processing community,  by characterizing different methods for training and constructing GANs, and challenges in the theory and applications.
In \cite{ghosh2020ICCCNT} Ghosh et al. presented a comprehensive summary of the progression and performance of GANs along with their various applications.
Saxena et al. \cite{saxena2020arXiv} conducted a survey of the advancements in GANs design, and optimization solutions proposed to handle GANs challenges.
Kumar et al. \cite{kumar2020Springer} presented state-of-the-art research on GANs, their applications, evaluation metrics, challenges, and benchmark datasets. 
In \cite{salehi2020arXiv} two new deep generative models, including GA, were compared, and the most remarkable GAN architectures were categorized and discussed.

Gui et al. in \cite{gui2020arXiv} provided a review of various GANs methods from an algorithmic perspective, theory, and applications. Jabbar et al.~\cite{jabbar2020arXiv} reviewed different GAN variants, applications, and several training solutions.
Hitawala in \cite{hitawala2018arXiv} presented different versions of GANs and provided a comparison between them from some aspects, such as learning, architecture, gradient updates, object, and performance metrics.
In a similar manner, Gonog et al. in \cite{gonog2019ICIEA} reviewed the extensional variants of GANs, and classified them in terms how they optimized the original GAN or changeed its basic structure, as well as their learning methods.
In \cite{hong2019CSUR} Hong et al. discussed the details of the GAN from the perspective of various object functions, architectures, and the theoretical and practical issues in training the GANs.  The authors also enumerate the GAN variants applied in different domains.
Bissoto et al. in \cite{bissoto2019arXiv} conducted a review of GAN advancements in six fronts, including architectural contributions, conditional techniques, normalization and constraint contributions, loss functions, image-to-image translations, and validation metrics.
Zhang et al. \cite{zhang2018IEEE} surveyed twelve extended GAN models and classified them in terms of the number of game players.
Pan et al.  in \cite{pan2019IEEE} analyzed the differences among different generative models, and classified them from the perspective of architecture and objective function optimization. They also discussed the training tricks and evaluation metrics, and presented GANs applications and challenges.
\subsection{GAN Challenges}
In the second group of surveys, Lucic in \cite{lucic2017gans} conducted an empirical comparison of GAN models, with focus on unconditional variants.  
Alqahtani et al. \cite{alqahtani2019Springer} focused on potential applications of GANs in different domains. They attempted to identify advantages, disadvantages and major challenges for successful implementation of GANs.
Wiatrak et al. \cite{wiatrak2019arXiv} surveyed current approaches for stabilizing the GAN training procedure, and categorizing various techniques and key concepts.
In \cite{lee2020arXiv}, Lee et al. reviewed the regularization methods used in the stable training of GANs, and classified them into several groups by their operation principles.
In contrast, \cite{pan2020IEEE} performed a survey for the loss functions used in GANs, and analyzed the pros and cons of these functions.
As differentially private GAN models provides a promising direction for generating private synthetic data, Fan et al. in \cite{fan2020AAAI} surveyed the existing approaches presented for this purpose.
\subsection{GAN Applications}
The authors in \cite{cao2018IEEE, wang2019arXiv, wu2017TUP, wang2020IEEEAccess, sorin2020Elsevier, tschuchnig2020Patterns, agnese2020Wiley, jain2020JCR, sampath2020, yi2019Elsevier} conducted reviews of different aspects of GAN progress in the field of computer vision and image synthesis.
Cao et al. \cite{cao2018IEEE} reviewed recently GAN models and their applications in computer vision. Cao et al. in \cite{cao2018IEEE} compared the classical and stare-of-the art GAN algorithms  in terms of the mechanism, visual results of generated samples, and so on.
Wang et al. \cite{wang2019arXiv} structured a review towards addressing practical challenges relevant to computer vision. They discussed the most popular architecture-variant, and loss-variant GANs, for tackling these challenges.
Wu et al. in \cite{wu2017TUP} presented a survey of image synthesis and editing, and video generation with GANs. They covered recent papers that leverage GANs in image applications, including texture synthesis, image inpainting, image-to-image translation, image editing, as well as video generation.
Along the same lines, Wang et al.~\cite{wang2020IEEEAccess} introduced the recent research on GANs in the field of image processing, and categorized them in four fields including image synthesis, image-to-image translation, image editing, and cartoon generation.

Surveys such as \cite{agnese2020Wiley} and \cite{jain2020JCR} focused on reviewing recent techniques to incorporate GANs in the problem of text-to-image synthesis. In \cite{agnese2020Wiley}, Agnese et al. proposed a taxonomy to summarize GAN based text-to-image synthesis papers into four major categories: Semantic Enhancement GANs, Resolution Enhancement GANs, Diversity Enhancement GANS, and Motion Enhancement GANs.
Different from the other surveys in this field, Sampath et al. \cite{sampath2020} examined the most recent developments of GANs techniques for addressing imbalance problems in image data. The real-world challenges and implementations of synthetic image generation based on GANs covered in this survey.

In \cite{yi2019Elsevier, sorin2020Elsevier, tschuchnig2020Patterns}, the authors dealt with the medical applications of image synthesis by GANs. Yi et al. in \cite{yi2019Elsevier} described the promising applications of GANs in medical imaging, and identifies some remaining challenges that need to be solved. As another paper in this subject, \cite{sorin2020Elsevier} reviewed GANs' application in image denoising and reconstruction in radiology. Tschuchnig et al. in \cite{tschuchnig2020Patterns} summarized existing GAN architectures in the field of histological image analysis.

The authors of \cite{yinka2019Springer} and \cite{ ghosh2020ICCCNT} provided reviews on the GANs in the cybersecurity. Yinka et al. \cite{yinka2019Springer} surveyed studies where the GAN plays a key role in the design of a security system or adversarial system.
Ghosh et al. \cite{ghosh2020ICCCNT} focused on the various ways in which GANs have been used to provide both security advances and attack scenarios in order to bypass detection systems. 

Di Mattia et al. \cite{di2019arXiv} surveyed the principal GAN-based anomaly detection methods.
Geogres et al. \cite{geogres2020} reviewed the published literature on Observational Health Data to uncover the reasons for the slow adoption of GANs for this subject.
Gao et al. in \cite{gao2020arXiv} addressed the practical applications and challenges relevant to spatio-temporal applications, such as trajectory prediction, events generation and time-series data imputation.
The recently proposed user mobility synthesis schemes based on GANs summarized in \cite{shin2020ICACT}.

According to the classification provided for review papers, our survey falls into the second category. We focus specifically on the recent progress of the application of game-theoretical approaches towards addressing the GAN challenges. While several surveys for GANs have been presented to date, to the best of our knowledge, our survey is the first to address this topic. Although the authors in \cite{wiatrak2019arXiv} presented a few game-model GANs, they have not done a comprehensive survey in this field, and many new pieces of research have not been covered. We hope that our survey will serve as a reference for interested researchers on this subject, who would like to extend GANs with leverage on game theoretical approaches.


\definecolor{LightGray}{gray}{0.9}
\definecolor{LightBlue}{rgb}{0.8,0.87,0.95}
\section{ Game of GANs: A Taxonomy}
\label{Sec:Taxonomy}
\begin{figure*}[t]
	\centering	
	\begin{tikzpicture}
	{\small
		\node[box,anchor=south, fill=yello]{Proposed GAN Taxonomy}
		[edge from parent fork down]
		child	{node[box, fill=orang]{Modified Game Model}
			child[grandchild,first]
			{node[box]{Stochastic game \cite{franci2020arXiv} }}
			child[grandchild,second]
			{node[box]{Stackelberg game \cite{zhang2018arXiv, farnia202arXiv} }}
			child[grandchild,third]
			{node[box]{Bi-affine game \cite{hsieh2019PMLR} }}
		}
		child	{node[box, fill=bluee]{Modified Learning Method}
			child[grandchild,first]
			{node[box]{No regret learning \cite{kodali2017arXiv, goodfellow2016arXiv, grnarova2017arXiv} }}
			child[grandchild,second]
			{node[box]{Fictitious play \cite{ge2018ECCV}}}
			child[grandchild,third]
			{node[box]{Federated learning \cite{rasouli2020arXiv, fan2020arXiv}}}
			child[grandchild,forth]
			{node[box]{Reinforcement learning \cite{zhang2018cybernetics, florensa2018PMLR, yu2017AAAI, xu2018EMNLP, yan2018ICPR, de2018arXiv, guimaraes2017arXiv, li2018Elsevier, aghakhani2018SPW, ghosh2016arXivHandwriting, hossam2020arXiv, tian2020Springer}}}
		}
		child	{node[box, fill=grenn]{Modified Architecture}
			child[grandchild,first]
			{node[box,big]{Multiple generators, One discriminator \cite{zhang2018arXiv, ghosh2018IEEE, hoang2018ICLR, ke2020IEEE, ghosh2016arXiv} }}
			child[grandchild,second]
			{node[box,big]{One generator, Multiple discriminators \cite{durugkar2016arXiv, jin2020ICME, hardy2019IPDPS, aghakhani2018SPW, mordido2020IEEE, nguyen2017ANIPS} }}
			child[grandchild,third]
			{node[box,big]{Multiple generators, Multiple discriminators \cite{arora2017arXiv, rasouli2020arXiv, ke2020IEEE} }}
			child[grandchild,forth]	
			{node[box,big]{One generator, One discriminator, One classifier \cite{tran2019ANIPS, li2018Elsevier}}}
			child[grandchild,fifth]
			{node[box,big]{One generator, One discriminator, One RL agent \cite{de2018arXiv, sarmad2019CVPR, liang2019IEEE, guimaraes2017arXiv} }}
		};}
	\end{tikzpicture}
	\caption{Proposed taxonomy of GAN advances by game theory.}
	\label{Fig:Taxonomy}
\end{figure*}
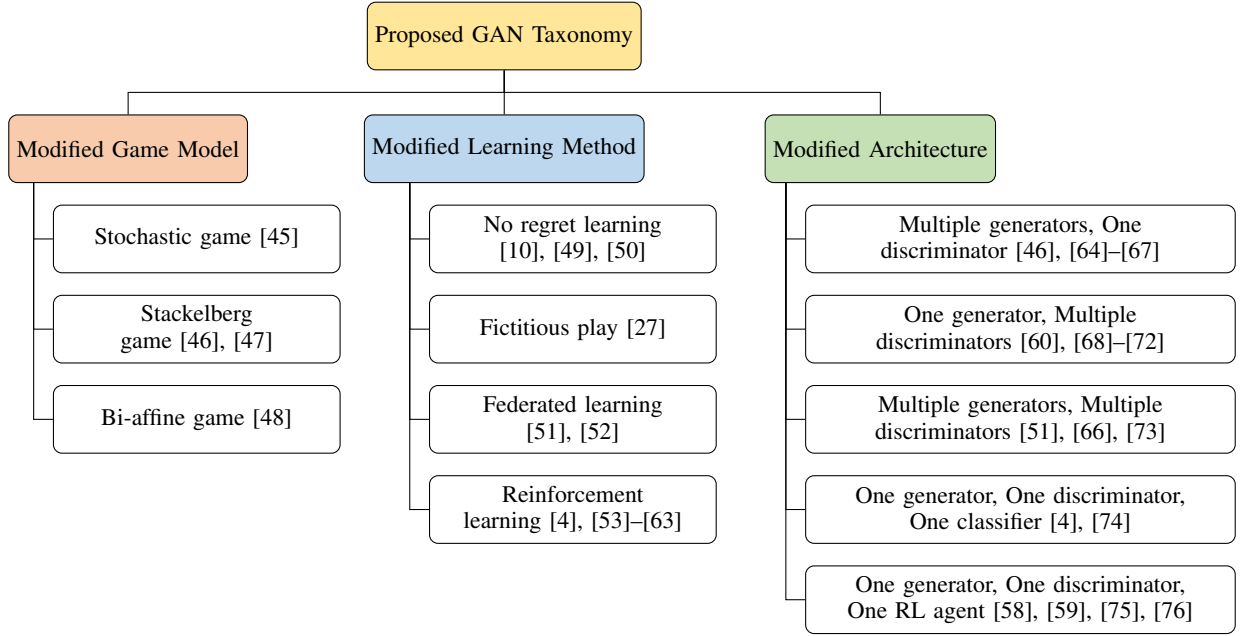
In this section, we present taxonomy for classifying GANs into three main categories, by focusing on how these works extended the original GAN. In the taxonomy is done in terms of (1) \emph{modified game mode}, (2) \emph{architecture modification}, and (3) \emph{modified learning algorithms}. Based on these there criteria, we further classify each category into several subsets, as shown in Fig. \ref{Fig:Taxonomy}. In the following sections, we introduce each category and the recent advances in each group will be discussed.
\subsection{Modified Game Model}
\label{Modified_Game_Model}
The core of all GANs is a competition between a generator and a discriminator, which model as a zero-sum game. Therefore, game theory plays a key role in this context. However, most of GANs relying on the basic model, formulating it as a two-player zero-sum (minimax) game, but some research utilized other game variants to tackle the challenges in this field. In this section, we aim to review these literatures. We classify the works under this category into three subcategories. Section \ref{SubSec:Stochastic_game} presents researches that cast the training process as a stochastic game. Research works presented in Section \ref{SubSec:Stackelberg_game} apply the idea of leader-follower of the Stackelberg game in the GANs. Finally Section \ref{SubSec:Bi-affine_game} presents GANs models as a Bi-affine game. A
summary of the reviewed researches in the modified game model category is shown in Table \ref{Tab:Summery_Modified_Game_Model}.
\subsubsection{Stochastic game}
\label{SubSec:Stochastic_game}

One of the main issues for GANs is that these neural networks are very hard to train because of the convergence problems. Franci et al. in \cite{franci2020arXiv} addressed this problem by casting the training procedure as a \emph{stochastic Nash equilibrium problem} (SNEP). The SNEP will recast as a \emph{stochastic variational inequality} (SVI) and target the solutions that are SNE. The advantage of this approach is that there are many algorithms for finding the solution of an SVI, like the forward-backward algorithm, also known as gradient descent. 
Franci et al. proposed a \emph{stochastic relaxed forward-backward} (SRFB) algorithm and a variant with an additional step for averaging over decision variables (aSRFB) for the training process of GANs.
For proving convergence to a solution, we need monotonicity on the pseudogradient mapping, which is defined by Equation \eqref{eq:pseudogradient}, where $J_g$ and $J_d$ are the payoff functions of the generator and the discriminator.
\begin{equation}
	\mathbb{F} = \begin{bmatrix}
		\mathbb{E}[\nabla_{x_g} J_g(x_g, x_d)]\\
		\mathbb{E}[\nabla_{x_d} J_d(x_d, x_g)]
	\end{bmatrix}
	\label{eq:pseudogradient}
\end{equation}

If pseudogradient mapping of the game is monotone and the increasing number of samples is available, the algorithm converges to the exact solution but with only finite, fixed mini-batch samples, and by using the averaging technique, it will converge to a neighborhood of the solution.

\subsubsection{Stackelberg game}
\label{SubSec:Stackelberg_game}

\DeclarePairedDelimiter{\norm}{\lVert}{\rVert} 

One of the main issues for GAN is the convergence of the algorithm. Farnia et al. in \cite{farnia202arXiv} showed that "GAN zero-sum games may not have any local Nash equilibria" by presenting certain theoretical and numerical examples of standard GAN problems. Therefore, based on the natural sequential type of GANs, where the generator moves first and follows the discriminator (leader), this problem can be considered as a Stackelberg game and focused on \emph{subgame perfect equilibrium} (SPE). For solving the convergence issue, the authors tried to find the equilibrium called \emph{proximal equilibrium} which enables traversing the spectrum between Stackelberg and Nash equilibria. In a proximal equilibrium, as shown in Equation~\eqref{eq:v_prox}, we allow the discriminator locally to optimize in a norm-ball nearby the primary discriminator. To keep the $\tilde{D}$ close to $D$, they penalize the distance among the two functions by $\lambda$, as $\lambda$ goes from zero to infinity, the equilibria change from Stackelberg to Nash.

\begin{equation}
	V_{\lambda}^{prox}(G,D) := \max_{\tilde{D} \in D} V(G, \tilde{D})- \frac{\lambda}{2}\norm{\tilde{D}-D}^2
	\label{eq:v_prox}
\end{equation}

Farnia et al. also proposed proximal training which optimizes the proximal objective $V_{\lambda}^{prox}(G, D)$ instead of the original objective $V(G, D)$ that can apply to any two-player GAN. Zhang et al. in \cite{zhang2018arXiv} also this game and presented Stackelberg GAN to tackles the instability of GANs training process. Stackelberg GAN is using a multi-generator architecture and the competition is between the generators (followers) and the discriminator (leader). We discussed the architecture details in Section \ref{SubSec:Multiple_generators_One_discriminator}.

\subsubsection{Bi-affine game}
\label{SubSec:Bi-affine_game}

Hssieh et al. in \cite{hsieh2019PMLR} examined training of GANs by reconsidering the problem formulation from the mixed NE perspective. In the absence of convexity, the theory focuses only on the local convergence, and it implies that even the local theory can break down if intuitions are blindly applied from convex optimization. In \cite{hsieh2019PMLR} the mixed Nash Equilibria of GANs are proposed. They are global optima of infinite-dimensional bi-affine games. Finite-dimensional bi-affine games are also applied for finding mixed NE of GANs. They also show that we can relax all current GAN objectives into their mixed strategy forms. Eventually, in this article, it’s experimentally shown that their method achieves better or comparable performance than popular baselines such as SGD, Adam, and RMSProp.

\begin{table*}[t]
   \centering
   \caption{Summary of publications included in the modified game model category (Section \ref{Modified_Game_Model})}
   \setlength{\tabcolsep}{4pt}
   \begin{tabularx}{\textwidth}{|p{1.61cm}|p{1.8cm}|p{4.6cm}|p{4cm}|X|} 
   \hline
      \rowcolor{LightBlue}
	  Reference & Convergence & Methodology and Contributions & Pros & Cons \\ 
	  \hline 
	  \rowcolor{LightGray}
	  \multicolumn{5}{|c|}{Stackelberg Game: Subsection \ref{SubSec:Stackelberg_game} }   \\  \hline
	   Stackelberg GAN \cite{zhang2018arXiv}
	   & SPNE
	   & Models multi-generator GANs as a Stackelberg game
	   & Can be built on top of standard GANs \& Proved the convergence
	   & - \\ 
	   \hline
	   \cite{farnia202arXiv}
	   & SPE
	   & Theoretical examples of standard GANs with no NE. \& proximal equilibrium as a solution, proximal training for GANs
	   & Can apply to any two-player GANs, allow the discriminator to locally optimize
	   & Focus only on the pure strategies of zero-sum GANs in non-realizable settings\\ 
	   \hline
	  \rowcolor{LightGray}
	  \multicolumn{5}{|c|}{Stochastic Game: Subsection \ref{SubSec:Stochastic_game} }   \\  \hline
      \cite{franci2020arXiv}& SNE & Cast the problem as SNEP and recast it to SVI, SRFB and aSRFB solutions & Proved the convergence & Need monotonicity on the pseudogradient mapping, increasing number of samples to reach an equilibrium \\ \hline
	 \rowcolor{LightGray}
	 \multicolumn{5}{|c|}{Bi-affine Game: Subsection \ref{SubSec:Bi-affine_game} }   \\  \hline
     \cite{hsieh2019PMLR}& Mixed NE &Tackling the training of GANs by reconsidering the problem formulation from the mixed Nash Equilibria perspective &Showing that all GANs can be relaxed to mixed strategy forms, flexibility & - \\ \hline
	\end{tabularx}
\label{Tab:Summery_Modified_Game_Model}
\end{table*}
\subsection{Modified Architecture}
\label{Modified_Architecture}
As we mentioned in Section \ref{Sec:Background}, GAN is a framework for producing a generative model through a two-player minimax game; however, in recent works, by extending the idea of using a single pair of generator and  discriminator to the multi-agent setting, the two-player game transforms to multiple games or multi-agent games.

 In this section, we review literature in which proposed GAN variants have modified the architecture in such a way that we have GANs with a mixture of generators, and/or discriminators and show how applying such methods can provide better convergence properties and prevent mode collapse. The majority of the works in this category focuses on introducing a larger number of generators and/or discriminators. But, in some papers, the number of generators and discriminators did not change, and a new agent has been added converting the problem to a multi-agent scenario. 
 
In Section \ref{SubSec:Multiple_generators_One_discriminator}, we will discuss GAN variants which extended the basic structure from a single generator to many generators.
In Section \ref{SubSec:One_generator_Multiple_discriminators}, we are going to review articles that deal with the problem of mode collapse by increasing the number of discriminators in order to force the generator to produce different modes. Section \ref{SubSec:Multiple_generators_Multiple_discriminators} is dedicated to discussing papers which develop GANs with multiple generators and multiple discriminators. 
Articles will be reviewed in Sections \ref{SubSec:One_generator_One_discriminator_One_classifier} and \ref{SubSec:One_generator_One_discriminator_One_RL_agent} extend the architecture by adding another agent, which is a classifier (Section \ref{SubSec:One_generator_One_discriminator_One_classifier} ) or an RL agent (Section \ref{SubSec:One_generator_One_discriminator_One_RL_agent}), to show the benefits of adding these agents to GANs. The methodologies, contributions as well as the pros and cons of reviewed papers are summarized in Table \ref{Tab:Summery_Modified_Architecture}.

\subsubsection{Multiple generators, One discriminator}
\label{SubSec:Multiple_generators_One_discriminator}

The minimax gap is smaller in GANs with multi-generator architecture and more stable training performances are experienced in these GANs \cite{zhang2018arXiv}. As we mentioned in Section \ref{SubSec:Stackelberg_game}, Zhang et al. in \cite{zhang2018arXiv} tackled the problem of instability during the GAN training as a result of a gap between minimax and maximin objective values. To mitigate this issue, they designed a multi-generator architecture and model the competition among agents as a Stackelberg game.
Results have shown the minimax duality gap decreases as the number of generators increases.
In this article, the mode collapse issue is also investigated and showed that this architecture effectively alleviates the mode collapse issue.
One of the significant advantages of this architecture is that it can be applied to all variants of GANs, e.g., Wasserstein GAN, vanilla GAN, etc.
Additionally, with an extra condition on the expressive power of generators, it is shown that Stackelberg GAN can achieve $\epsilon$-approximate equilibrium with $\tilde{\mathcal{O}}(1 / \epsilon)$ generator \cite{zhang2018arXiv}.

Furthermore, Ghosh et al. in \cite{ghosh2018IEEE} proposed a multi-generator and single discriminator architecture for GANs named Multi-Agent Diverse Generative Adversarial Networks (MAD-GAN). 
In this paper, different generators capture varied, high probability modes, and the discriminator is designed such that, along with finding the real and fake samples, identifies the generator that generated the given fake sample \cite{ghosh2018IEEE}. It is shown that at convergence, the global optimum value of $-(k+1) \log (k+1)+k \log k$ is achieved, where k is the number of generators.

Comparing presented models in \cite{ghosh2018IEEE} and \cite{zhang2018arXiv}, in MAD-GAN \cite{ghosh2018IEEE} multiple generators are combined with the assumption that the generators and the discriminator have infinite capacity, but in the Stackelberg GAN \cite{zhang2018arXiv} there is no assumption on the model capacity. Also, in MAD-GAN \cite{ghosh2018IEEE} the generators share common network parameters, although, in the Stackelberg GAN \cite{zhang2018arXiv} various sampling schemes beyond the mixture model is allowed, and each generator has free parameters.

The assumption that increasing generators will cover the whole data space is not valid in practice. So Hoang et al. in \cite{hoang2018ICLR}, in contrast with \cite{ghosh2018IEEE}, approximated data distribution by forcing generators to capture a subset of data modes independently of those of others instead of forcing generators by separating their samples. Thus, they established a minimax formulation among a classifier, a discriminator, and a set of generators. The classifier determines which generator generates the sample by performing multi-class classification. Each generator is encouraged to generate data separable from those produced by other generators, because of the interaction between generators and the classifier. In this model, multiple generators create the samples. Then one of them will be randomly picked as the final output similar to the mechanism of a probabilistic mixture model. Therefore they theoretically proved that, at the equilibrium, the Jensen-Shannon divergence (JSD) between final output and the data distribution is minimal. In contrast, the JSD amongst generators' distributions is maximal, hence the mode collapse problem is effectively avoided.
Moreover, the computational cost that is added to the standard GAN is minimal in the suggested model by applying parameter sharing. The proposed model can efficiently scale to large-scale datasets as well.

Ke et al. \cite{ke2020IEEE} proposed a new architecture, named multiagent distributed GAN (MADGAN). In this framework, the discriminator is considered as a leader, and the generator is considered as a follower. Moreover, this proposal is based on the social group wisdom and the influence of the network structure on agents. MADGAN can have a multi-generator and multi-discriminator architecture (e.g., two discriminators and four generators) as well as multiple generators and single discriminator architecture, which is our discussed topic in this section. As one of the vital contributions of MADGAN, it can train multiple generators simultaneously, and the training results of all generators are consistent.

\begin{figure*}[t]
  \centering
    \begin{subfigure}[t]{\columnwidth}
           \centering
  \includegraphics[scale=0.20]{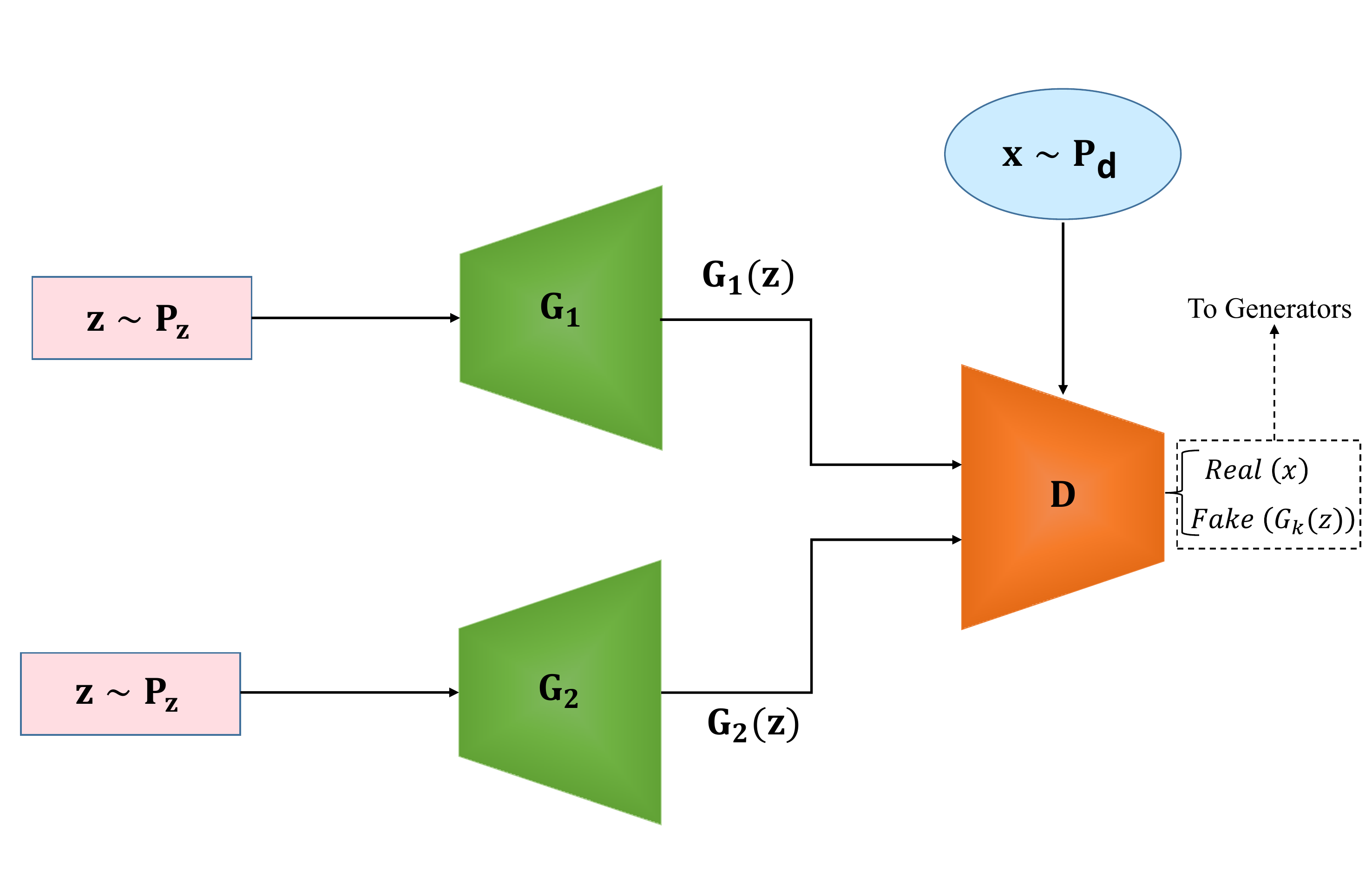}
           \caption{MADGAN \cite{ke2020IEEE}}
    \end{subfigure}
    \begin{subfigure}[t]{\columnwidth}
           \centering
  \includegraphics[scale=0.35]{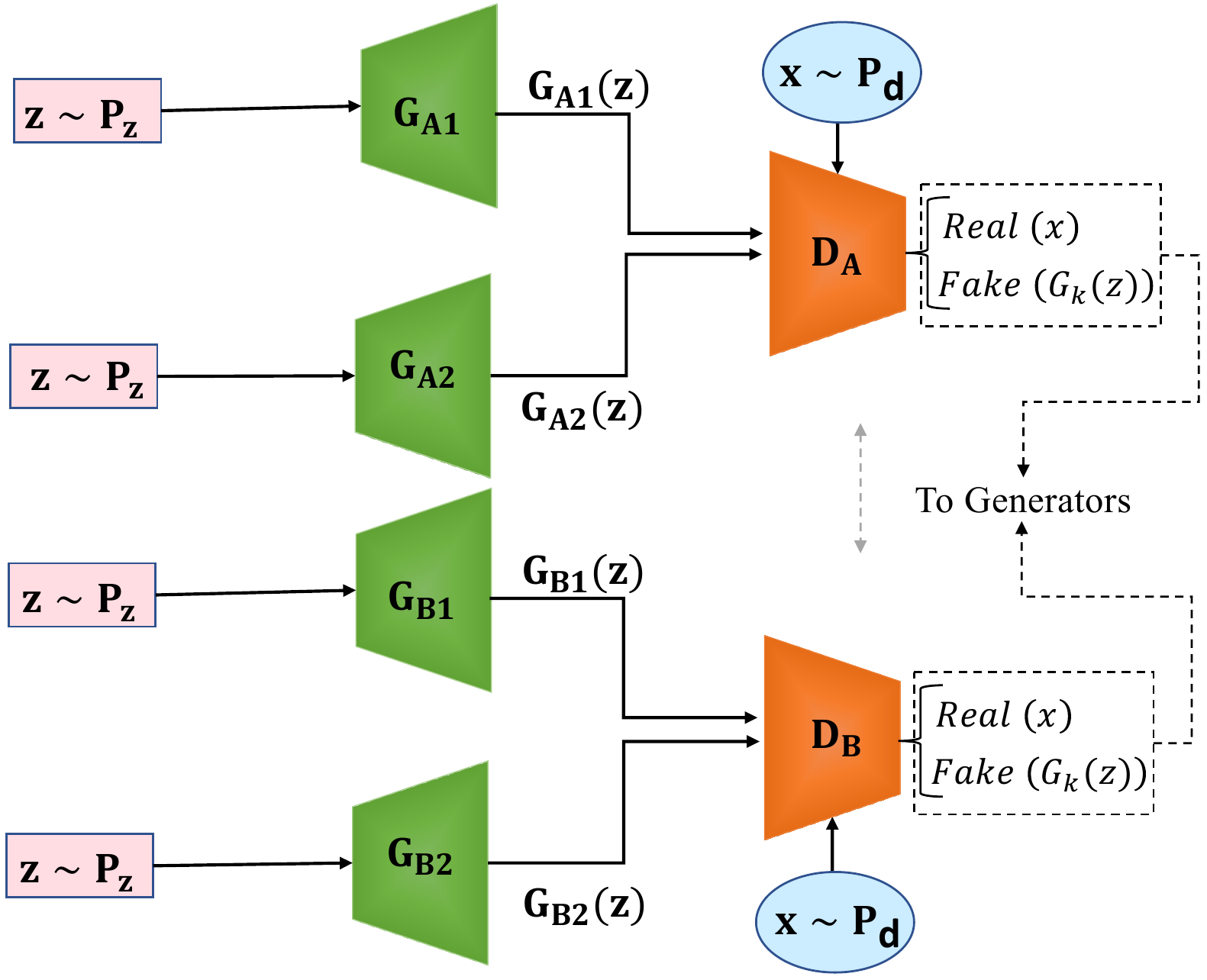}
           \caption{MADGAN \cite{ke2020IEEE} \\(with multiple discriminators, reviewed in Subsection~\ref{SubSec:Multiple_generators_Multiple_discriminators})}
    \end{subfigure}
    \begin{subfigure}[t]{\columnwidth}
           \centering
            \includegraphics[scale=0.25]{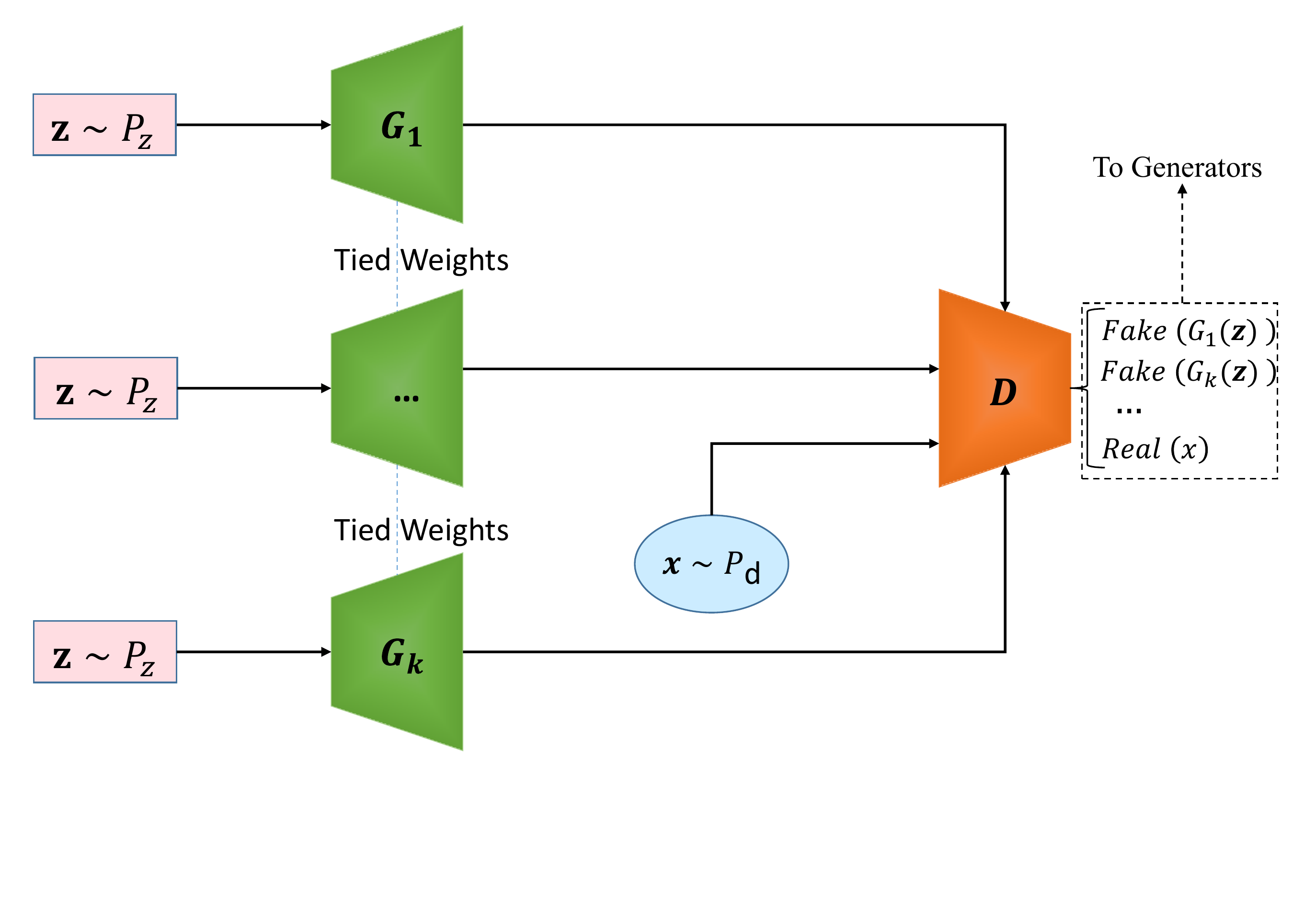}
           \caption{MAD-GAN \cite{ghosh2018IEEE}}
    \end{subfigure}
    \begin{subfigure}[t]{\columnwidth}
           \centering
            \includegraphics[scale=0.36]{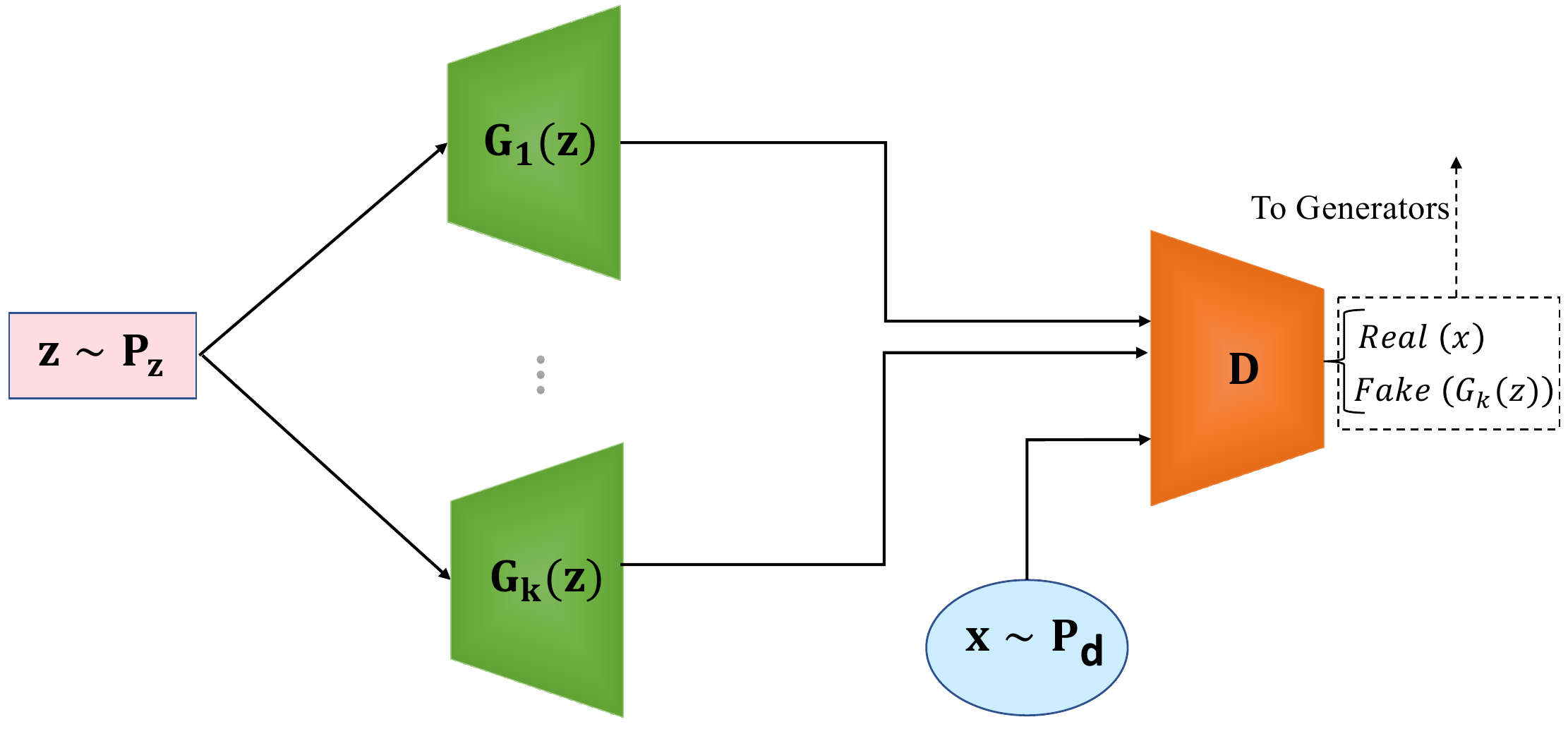}
           \caption{Stackelberg-GAN \cite{zhang2018arXiv}}
    \end{subfigure}
    \begin{subfigure}[t]{\columnwidth}
           \centering
            \includegraphics[scale=0.35]{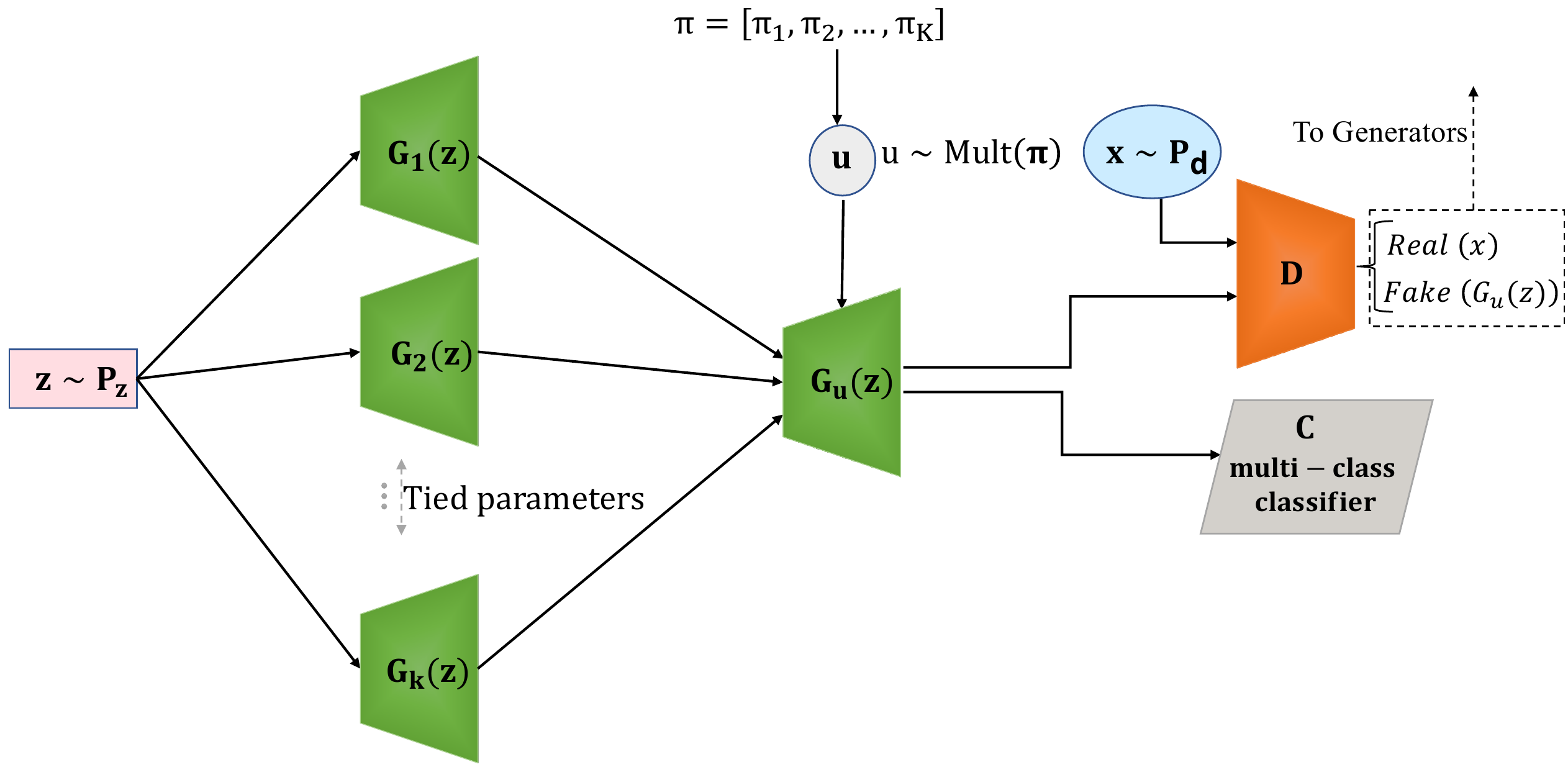}
           \caption{MGAN \cite{hoang2018ICLR}}
    \end{subfigure}
    \begin{subfigure}[t]{\columnwidth}
           \centering
            \includegraphics[scale=0.30]{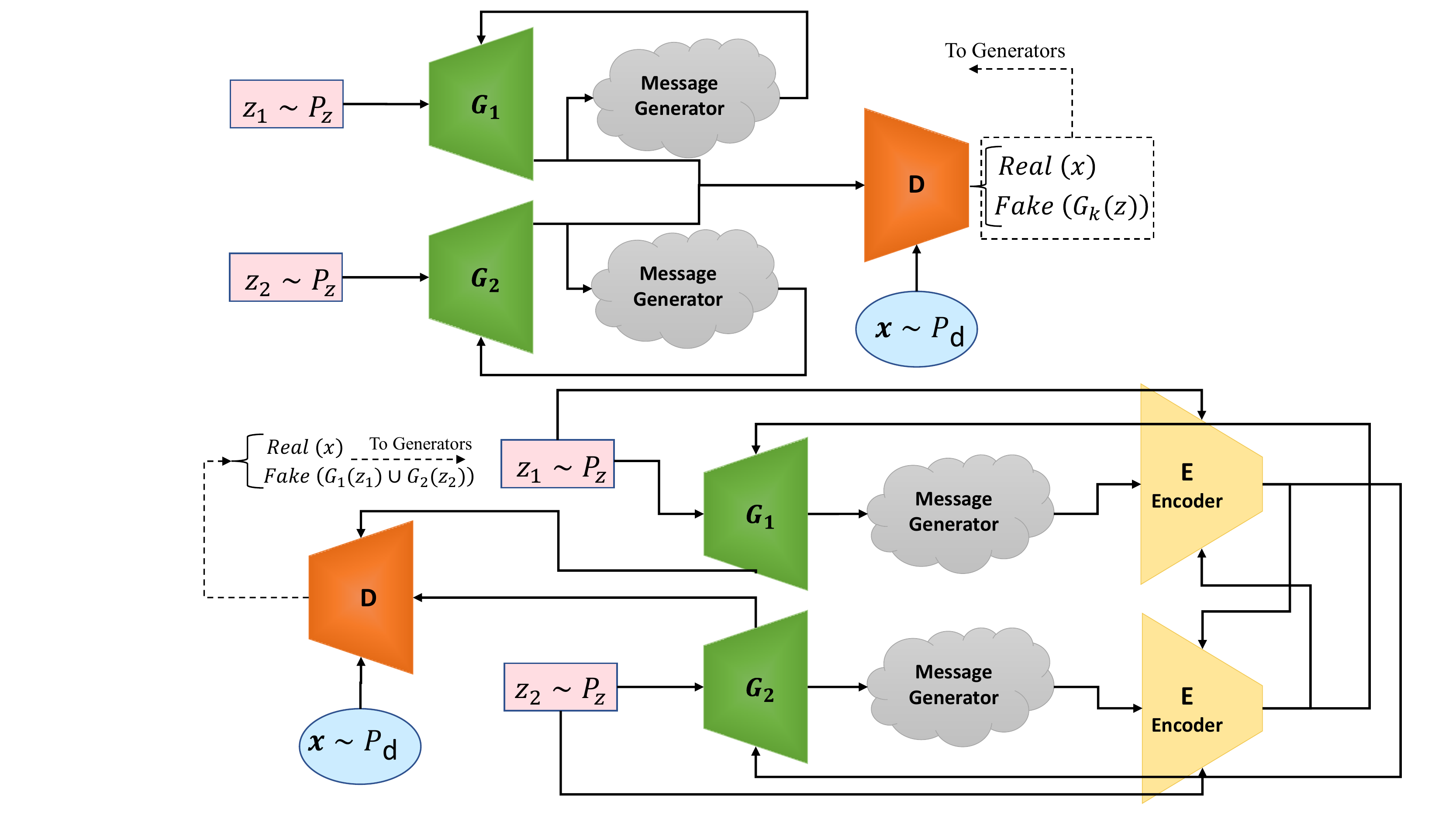}
           \caption{MPM-GAN \cite{ghosh2016arXiv}}
    \end{subfigure}
     \caption{A schematic view of GAN variants with multiple generators and one discriminator, presented in Subsection~\ref{SubSec:Multiple_generators_One_discriminator}. 
     D: Discriminator, G: Generator, $z$: Noise, $x$: Real data, $P_z$: Latent space, $P_x$: Data distribution, $G_i(z)$: Fake data generated by generator $i$, C: Classifier, E: Encoder.}
     \label{Fig:Architecures_Sec4.2.1}
\end{figure*}

Furthermore, in Message Passing Multi-Agent Generative Adversarial Networks \cite{ghosh2016arXiv}, the authors proposed that with two generators and one discriminator,  communicating through message passing, better image generation can be achieved. In this paper, there are two objectives such as competing and conceding. The competing introduced based on the fact that the generators compete with each other to get better scores for their generated data from the discriminator. However, the conceding introduced based on the fact that the two generators try to guide each other in order to get better scores for their generations from the discriminator and ensures that the message sharing mechanism guides the other generator to generate better than itself. Generally, in this section, innovative architectures and objectives aimed at training multi-agent GANs are presented. Fig.~\ref{Fig:Architecures_Sec4.2.1} presents a schematic view of GANs' structure reviewed in this section.


\subsubsection{One generator, Multiple discriminators}
\label{SubSec:One_generator_Multiple_discriminators}

The multi-discriminators are constructed with homogeneous network architecture and trained for the same task from the same training data. In addition to introducing a multi-discriminators schema, Durugkar et al. in \cite{durugkar2016arXiv}, from the perspective of game theory, showed that discriminators act like each other; thus, they will converge to similar decision boundaries. In the worst case, they may even converge to a single discriminator. So, Jin et al. in \cite{jin2020ICME} by discriminator discrepancy loss (DDL), the multiplayer minimax game unifies the optimization of DDL and the GAN loss, seeking an optimal trade-off between the accuracy and diversity of multi discriminators.
Compared to \cite{durugkar2016arXiv}, Hardy et al. in \cite{hardy2019IPDPS} distributed discriminators over multiple servers. Thus, they can train over datasets that are spread over numerous servers.

 Aghakhani et al. in \cite{aghakhani2018SPW} proposed FakeGAN. The GAN model uses two discriminators and one generator. The discriminators use the Monte Carlo search algorithm to evaluate and pass the intermediate action-value as the reinforcement learning (RL) reward to the generator. The generator is modeled as a stochastic policy agent in RL \cite{aghakhani2018SPW}.
Instead of one batch in \cite{jin2020ICME}, Mordido et al. in \cite{mordido2020IEEE} divided generated samples into multiple micro-batch. Then update each discriminator's task to discriminate between different samples. Samples coming from its assigned fake micro-batch and samples from the micro-batches assign to the other discriminator, together with the real samples.

Unlike \cite{durugkar2016arXiv}, Nguyen et al. in \cite{nguyen2017ANIPS} combined the Kullback-Leibler (KL) and reverse KL divergence (the measure of how one probability distribution is different from a second) into a unified objective function. Combining these two measures can exploit the divergence's complementary statistical properties to diversify the estimated density in capturing multi modes effectively.
From the perspective of game theory in \cite{nguyen2017ANIPS}, there are two discriminators and one generator with the analogy of a three-player minimax game. In this case, there exist two pair of players which are playing two minimax games simultaneously. In one of the games, the discriminator rewards high scores for samples from data distribution (i.e., reverse KL divergence, in Equation \eqref{eq:reverse_kl}), while another conversely rewards high scores for samples from the generator, and the generator produces data to fool two discriminators (i.e., KL divergence, in Equation \eqref{eq:kl}).
\begin{equation}
  \min\limits_{G}\max\limits_{D_1}  \jmath(G,D_1) = \alpha \times\sum_{x \sim P_{data}}[\log D_1(x)] + \sum_{z\sim P_z}[-D_1(G(z))],
  \label{eq:reverse_kl}
\end{equation}
\begin{equation}
  \min\limits_{G}\max\limits_{D_2} \jmath(G,D_2) = \sum_{x\sim P_{data}}[-D_2(x)] + \beta \times\sum_{z \sim P_{z}}[\log D_2(G(z))],
  \label{eq:kl}
\end{equation}
where in the above equations hyperparameters $\alpha , \beta$ are being used to control and stabilize the learning method.

\begin{figure*}[t]
  \centering
    \begin{subfigure}[t]{\columnwidth}
           \centering
  \includegraphics[scale=0.25]{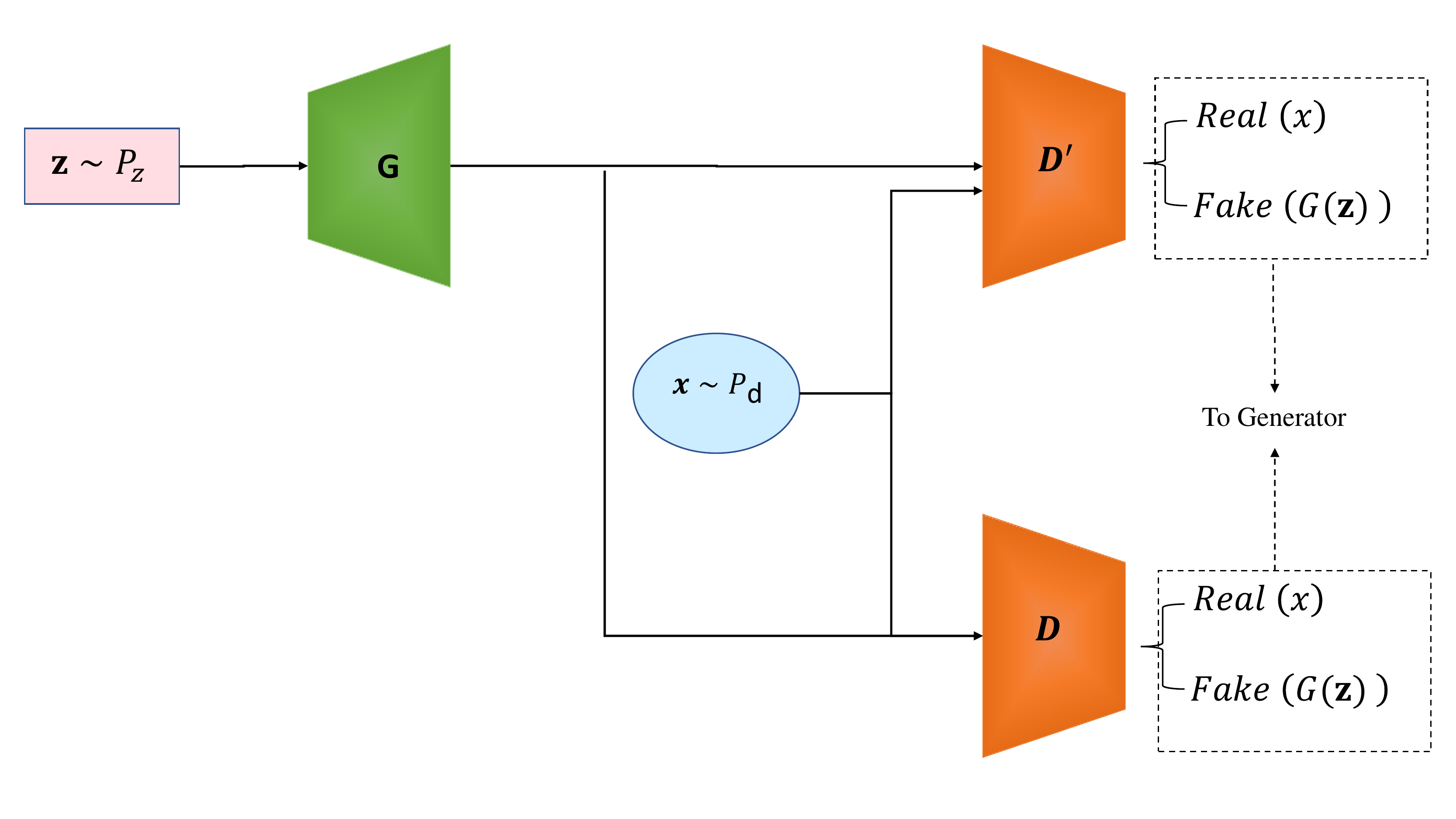}
           \caption{D2GAN \cite{nguyen2017ANIPS}}
    \end{subfigure}
    \begin{subfigure}[t]{\columnwidth}
           \centering
  \includegraphics[scale=0.25]{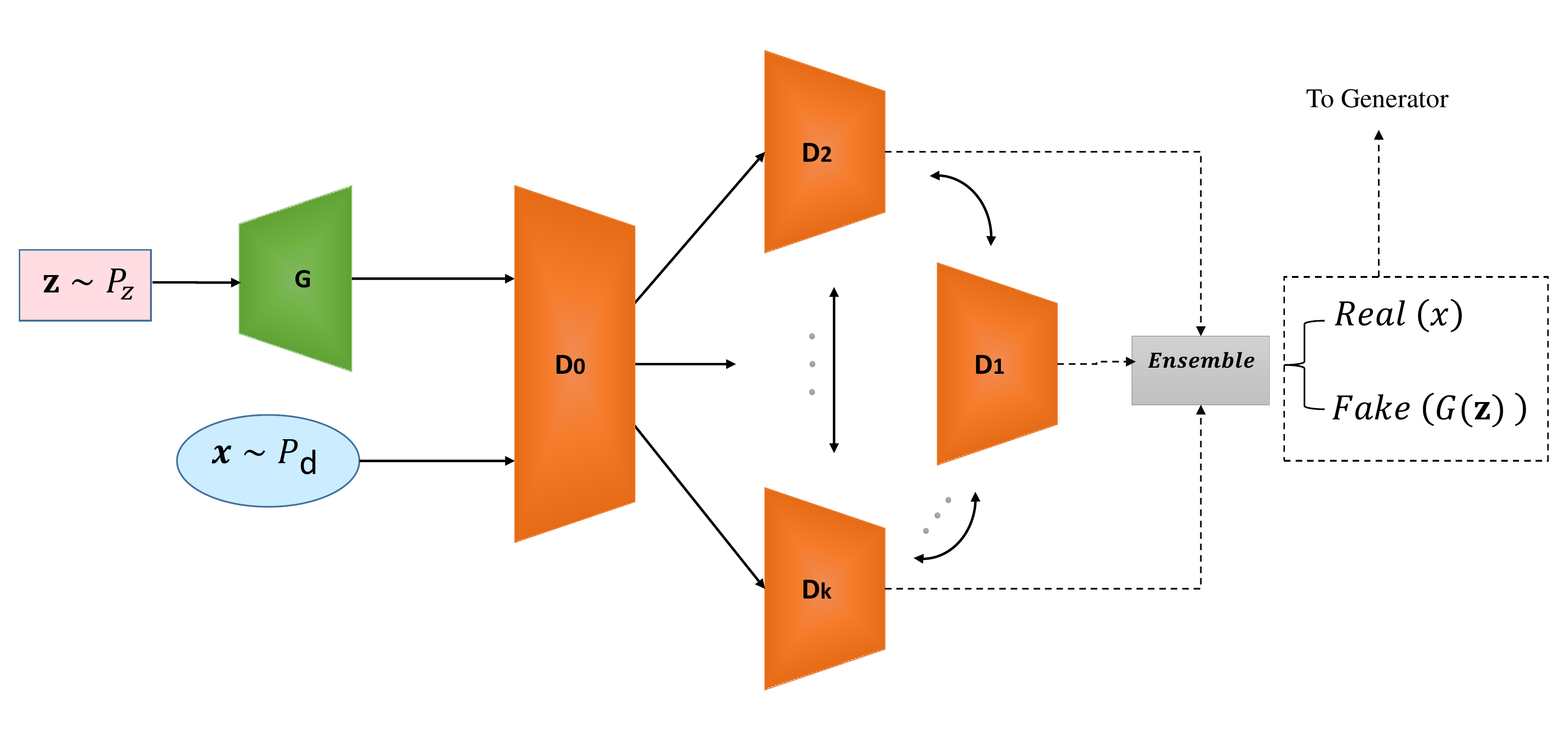}
           \caption{DDL-GAN \cite{jin2020ICME}}
    \end{subfigure}
    \begin{subfigure}[t]{\columnwidth}
           \centering
            \includegraphics[scale=0.25]{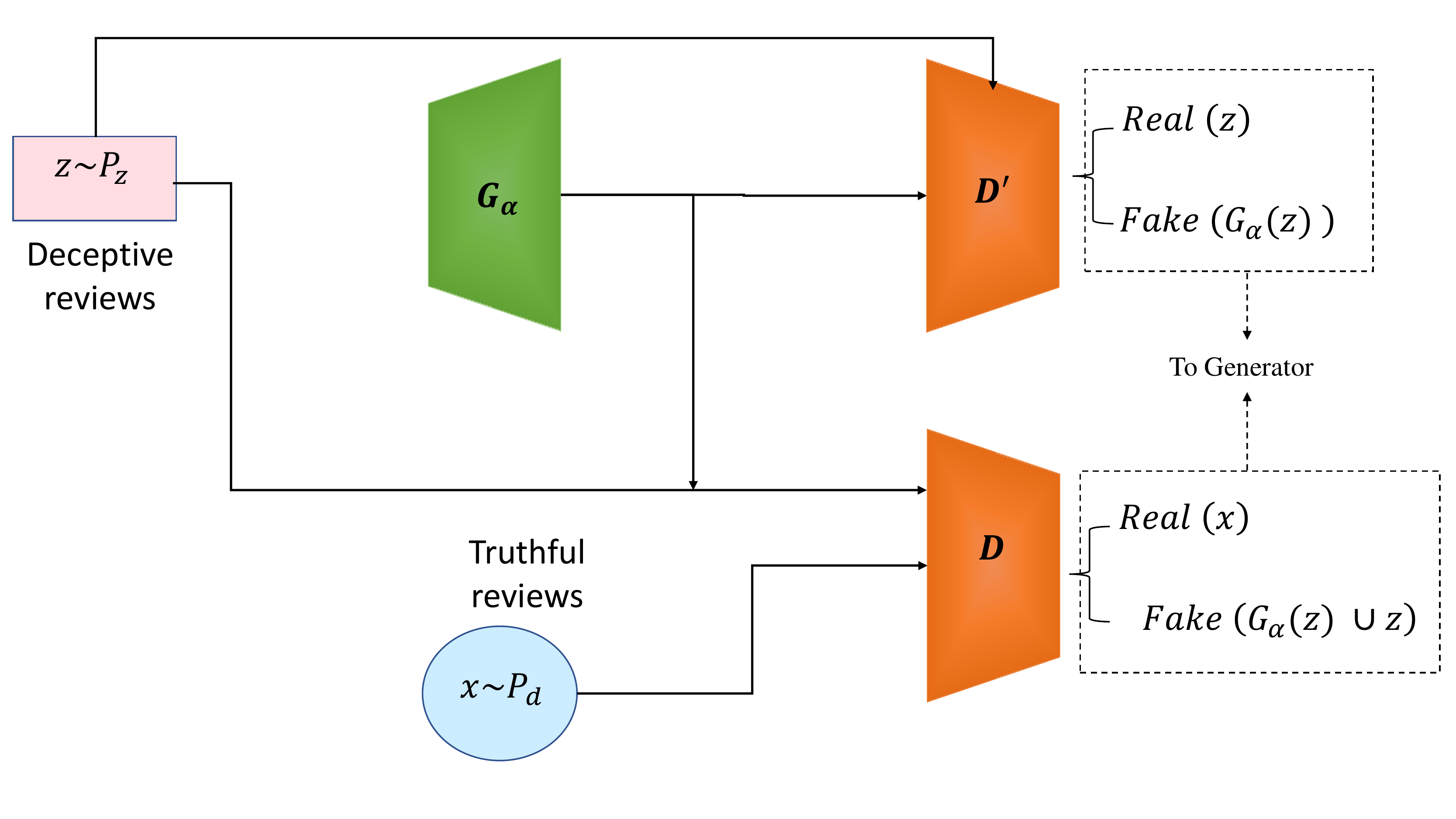}
           \caption{FakeGAN \cite{aghakhani2018SPW}}
    \end{subfigure}
    \begin{subfigure}[t]{\columnwidth}
           \centering
            \includegraphics[scale=0.25]{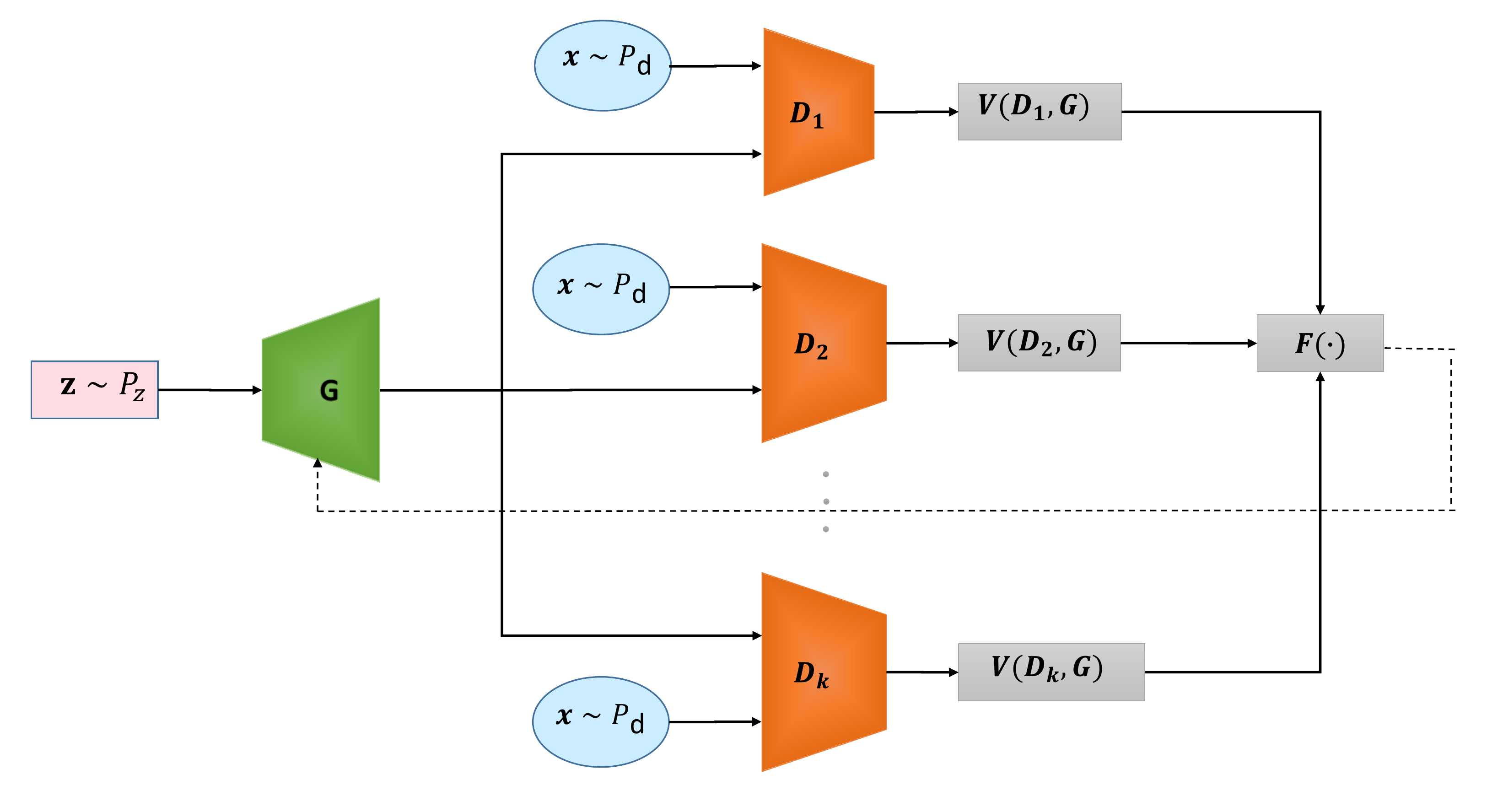}
           \caption{GMAN \cite{durugkar2016arXiv} (If F := max, G trains against the best discriminator. If F := mean, G trains against an ensemble)}
    \end{subfigure}
    \begin{subfigure}[t]{\columnwidth}
           \centering
            \includegraphics[scale=0.25]{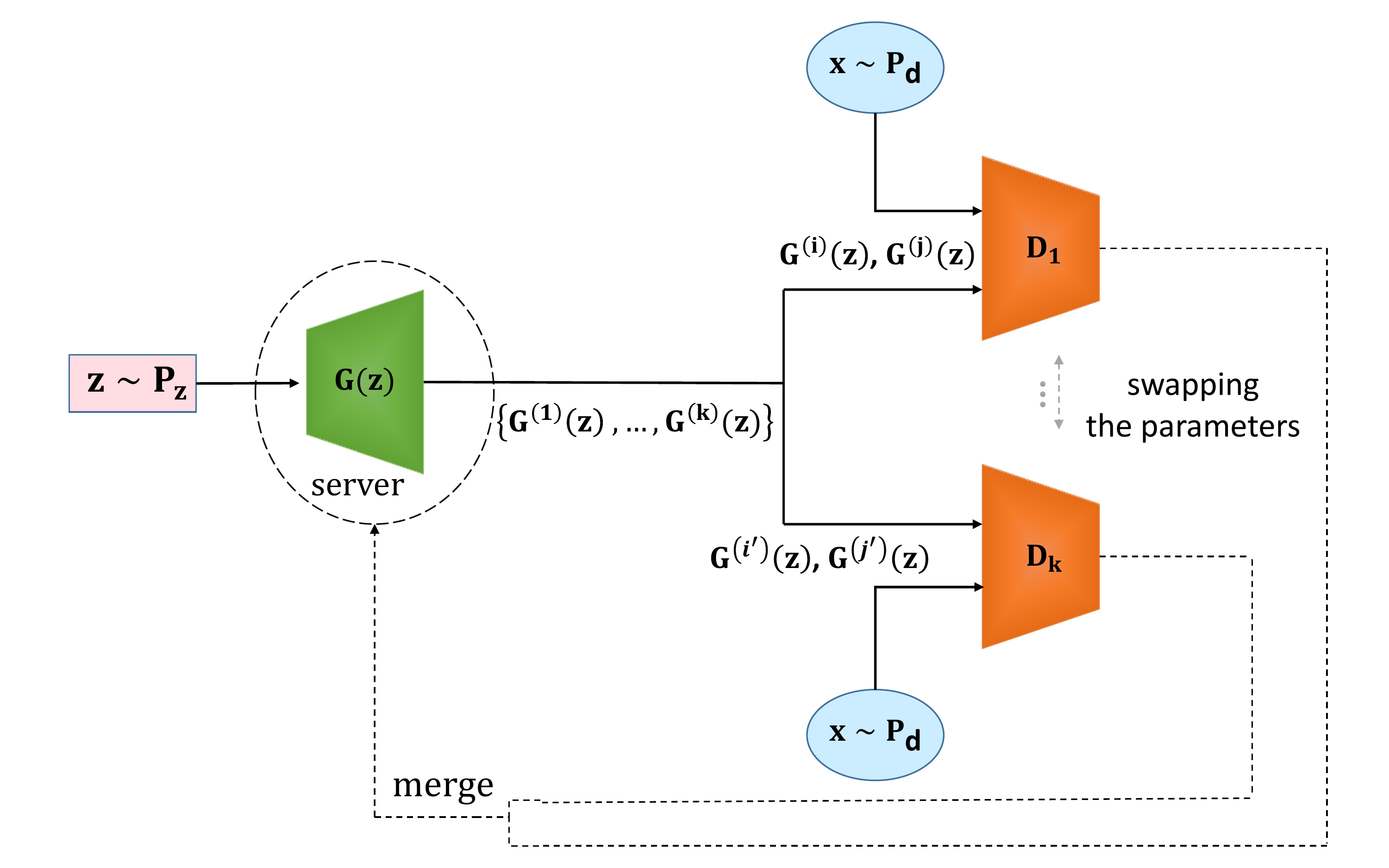}
           \caption{MD-GAN \cite{hardy2019IPDPS}}
    \end{subfigure}
    \begin{subfigure}[t]{\columnwidth}
           \centering
            \includegraphics[scale=0.25]{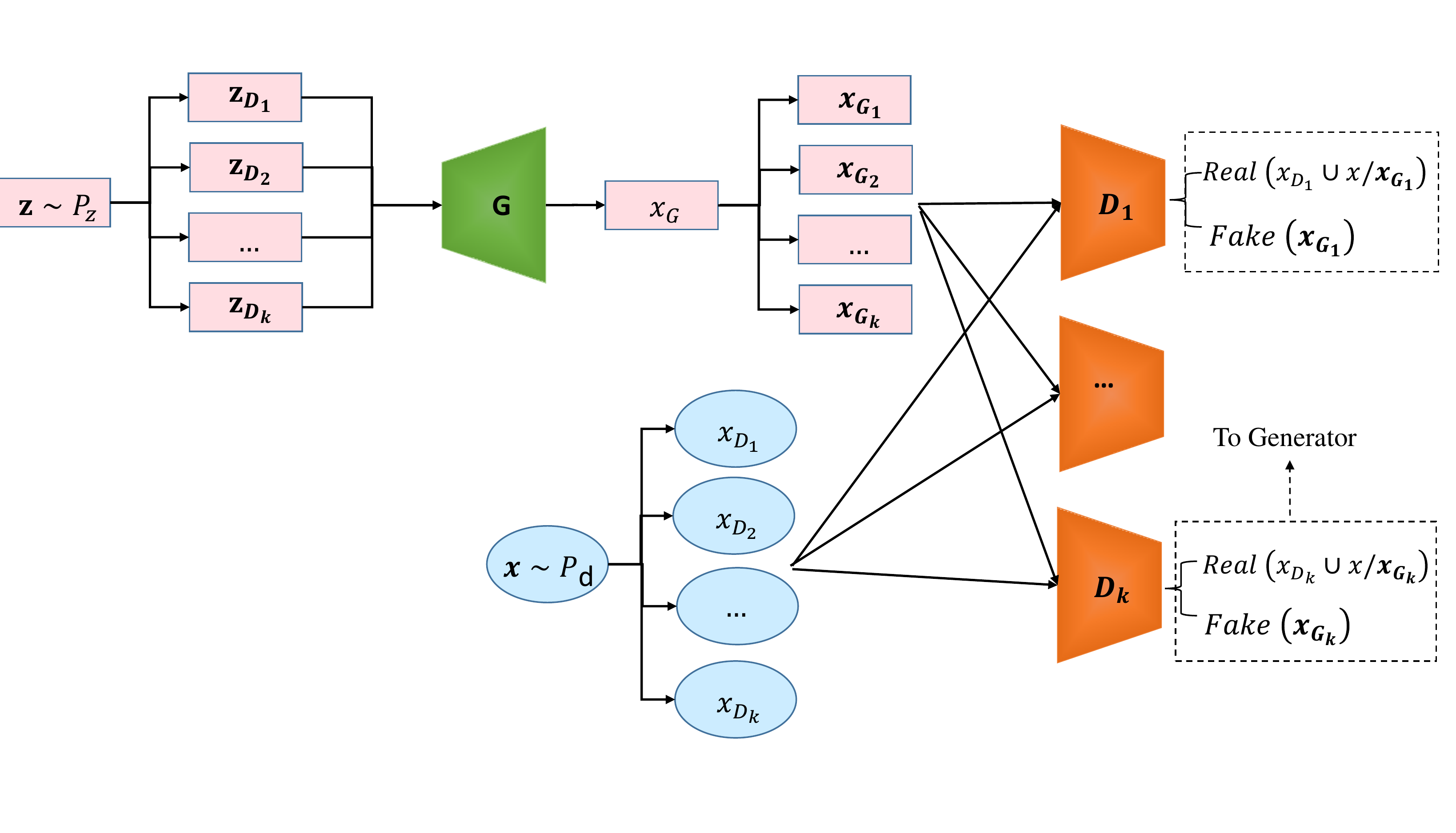}
           \caption{Microbatch-GAN \cite{mordido2020IEEE}}
    \end{subfigure}
     \caption{Schematic view of GANs with one generator and multiple discriminators, reviewed in Subsection~\ref{SubSec:One_generator_Multiple_discriminators}. 
     $D$: Discriminator, $G$: Generator, $z$: Noise, $x$: Real data, $P_z$: Latent space, $P_x$: Data distribution, $G(z)$: Fake data generated by the generator. }
     \label{Fig:Architecures_Sec4.2.2}
\end{figure*}

Minimizing the Kullback-Leibler (KL) divergence between data and model distributions covers multiple mods but may produce completely unseen and potentially undesirable samples.
In reverse KL divergence, it is observed that optimization towards the reverse KL divergence criteria mimics the mod seeking process where the $P_{model}$ concentrates on a single mode of $P_{data}$ while ignoring other modes. Fig.~\ref{Fig:Architecures_Sec4.2.2} presents a schematic view of the GANs reviewed in this section.


\subsubsection{Multiple generators, Multiple discriminators}
\label{SubSec:Multiple_generators_Multiple_discriminators}

The existence of equilibrium has always been considered one of the open theoretical problems in this game between generator and discriminator. Arora et al. in \cite{arora2017arXiv} turned to infinite mixtures of generator's deep nets in order to investigate the existence of equilibria. Unsurprisingly, equilibrium exists in an infinite mixture. Therefore, they showed that a mixture of a finite number of generators and discriminators can approximate min-max solution in GANs. This implies that an approximate equilibrium can be achieved with a mixture (not too many) of generators and discriminators. They also proposed a heuristic approximation to the mixture idea to introduce a new framework for training called MIX+GAN: use a mixture of T components, where T is as large as allowed by the size of GPU memory (usually T$\leq$5). In fact, a mixture of T generators and T discriminators are trained which share the same network architecture, but have their own trainable parameters. Maintaining a mixture represents maintaining a weight $w_{u_{i}}$ for the generator $G_{u_{i}}$ which corresponds to the probability of selecting the output of $G_{u_{i}}$ . These weights for the generator are updated by backpropagation. This heuristic can be applied to the existing methods like DCGAN, W-GAN, etc. Experiments show MIX+GAN protocol improves the quality of several existing GAN training methods and can lead to more stable training.

As we mentioned earlier, one of the significant challenges in GAN algorithms is their convergence. Refer to \cite{rasouli2020arXiv}, this challenge is a result of the fact that cost functions may not converge using gradient descent in the minimax game between the discriminator and the generator. Convergence is also one of the considerable challenges in federated learning. This problem becomes even more challenging when data at different sources are not independent and identically distributed. Therefore, Rasouli et al. proposed an algorithm for multi-generator and multi-discriminator architecture for training a GAN with distributed sources of non-independent-and-identically-distributed data sources named Federated Generative Adversarial Network (FedGAN)~\cite{rasouli2020arXiv}. Local generators and discriminators are used in this algorithm. These generators and discriminators are periodically synchronized via an intermediary that averages and broadcasts the generator and discriminator parameters. In fact, results from stochastic approximation for GAN convergence and communication-efficient SGD for federated learning are connected in this article to address FedGAN converge. One of the notable results in \cite{rasouli2020arXiv} is that FedGAN has similar performance to general distributed GAN while it converges and reduces communication complexity as well.

In \cite{ke2020IEEE}, Ke and Li proposed a multi-agent distributed GAN (MADGAN) framework based on the social group wisdom and the influence of the network structure on agents, in which the discriminator and the generator are regarded as the leader and the follower, respectively. The multi-agent cognitive consistency problem in the large-scale distributed network is addressed in MADGAN. In \cite{ke2020IEEE} the conditions of consensus presented for a multi-generator and multi-discriminator distributed GAN by analyzing the existence of stationary distribution to the Markov chain of multiple agent states. The experimental results show that the generation effect of the generators trained by MADGAN can be comparable to that of the generator trained by GAN. Moreover, MADGAN can train multiple generators simultaneously, and the training results of all generators are consistent.


\subsubsection{One generator, One discriminator, One classifier}
\label{SubSec:One_generator_One_discriminator_One_classifier}

One of the issues that GANs face is catastrophic forgetting in the discriminator neural network. Self-Supervised (SS) tasks were planned to handle this issue, however, these methods enable a seriously mode-collapsed generator to surpass the SS tasks. In \cite{tran2019ANIPS}, Tran et al. proposed new SS tasks, called  Multi-class minimax game based Self-supervised tasks (MS) which is based on a multi-class minimax game, including a discriminator, a generator, and a classifier. The SS task is a 4-way classification task of recognizing one among the four image rotations (i.e., $0$, $90$, $180$, and $270$ degrees). The discriminator SS task is to train the classifier C that predicts the rotation applied to the real samples and the generator SS task is to train the generator G to produce fake samples for maximizing classification performance. The SS task helps the generator learn the data distribution and generate diverse samples by closing the gap between supervised and unsupervised image classification. The theoretical and experimental analysis showed that the convergence of this approach has progressed.

\begin{figure*}[t]
  \centering
    \begin{subfigure}[t]{\columnwidth}
           \centering
            \includegraphics[scale=0.35]{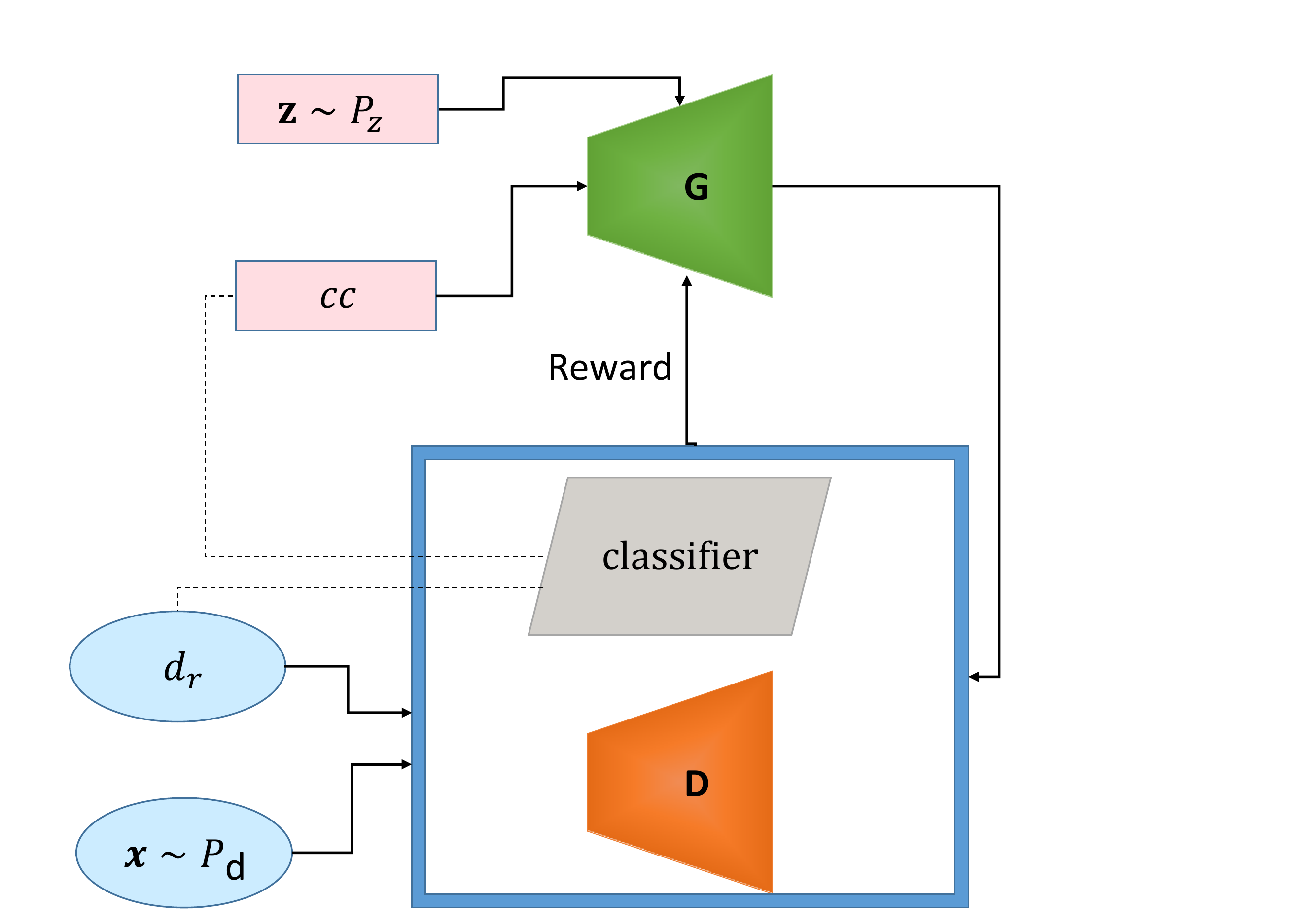}
           \caption{CS-GAN\cite{li2018Elsevier}}
    \end{subfigure}
    \begin{subfigure}[t]{\columnwidth}
           \centering
            \includegraphics[scale=0.35]{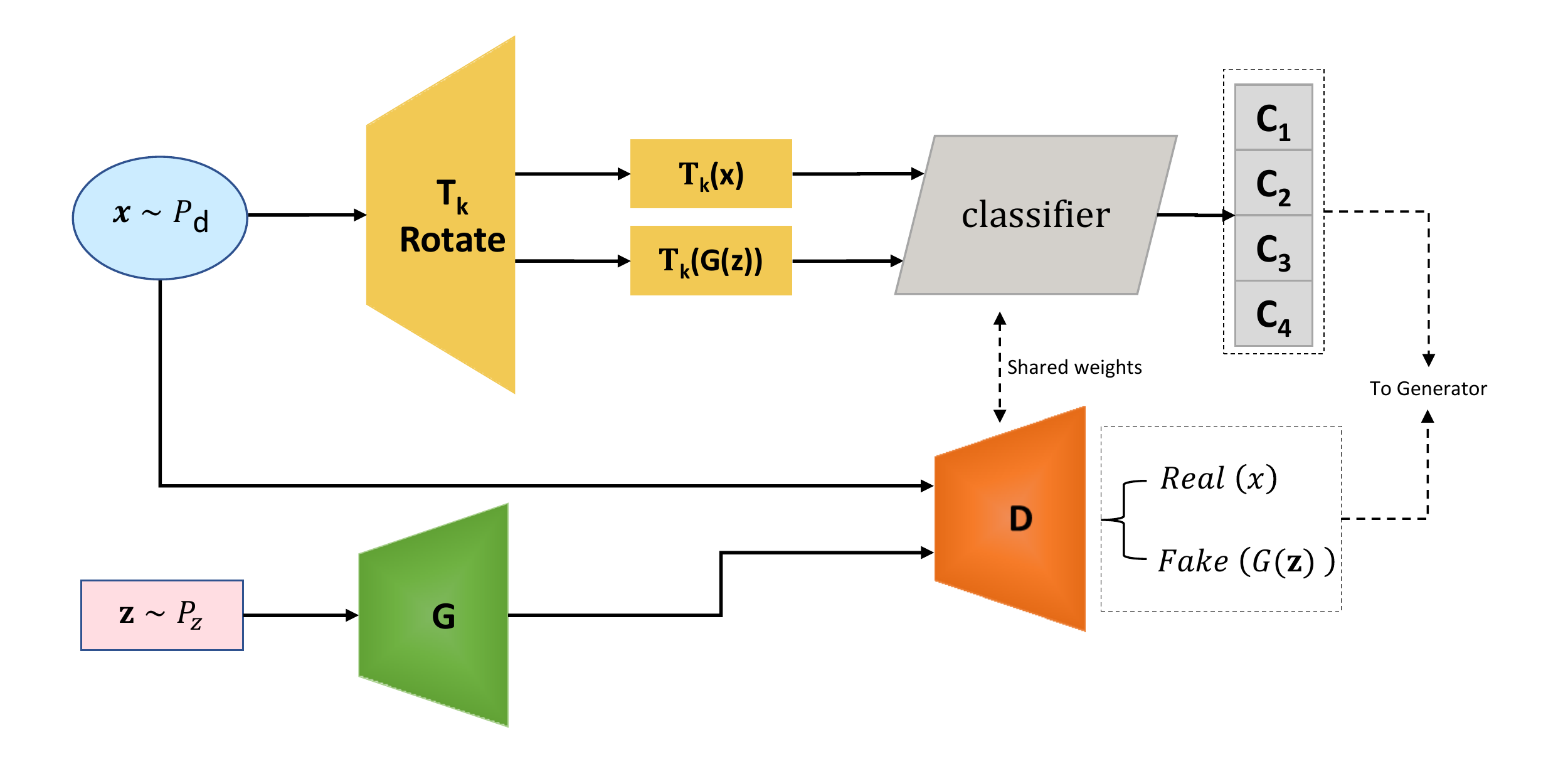}
           \caption{Self-supervised GAN \cite{tran2019ANIPS} $T_k, k=1,2,3,4$ are the transformations (0, 90, 180, 270 degree image rotation)
           }
    \end{subfigure}
     \caption{Schematic view of GANs with one generator, one discriminators, and one classifier reviewed in Subsection~\ref{SubSec:One_generator_One_discriminator_One_classifier}. 
     D: Discriminator, $G$: Generator, $z$: Noise, $x$: Real data, $P_z$: Latent space, $P_x$: Data distribution, $G(z)$: Fake data generated by the generator. }
     \label{Fig:Architecures_Sec4.2.4}
\end{figure*}

In~\cite{li2018Elsevier}, Li et al. also used a classifier generating categorized text. The authors proposed a new framework  Cyclic-Synthesized GAN (CS-GAN), which uses GAN, RNN, and RL to generate better sentences. The classifier position is to ensure that the generated text contains the label information and the RNN is a character predictor because the model is built at the character level to limit the large searching space.
We can divide the generation process into two steps, first adding category information into the model and making the model generate category sentences respectively, then combine category information in GAN to generate labeled sentences.
CS-GAN acts strongly in supervised learning, especially in the multi-categories datasets.
Fig.~\ref{Fig:Architecures_Sec4.2.4} presents architecture of GANs reviewed in this subsection.


\subsubsection{One generator, One discriminator, One RL agent}
\label{SubSec:One_generator_One_discriminator_One_RL_agent}

With an RL agent, we can deploy fast and robust control over the GAN's output or input. Moreover, such architecture can be used to optimize the generation process by adding an arbitrary (not necessarily differentiable) objective function to the model.

In \cite{de2018arXiv}, Cao et al. used this architecture for generating molecules and drug discovery. The authors encoded the molecules as the original graph-based representation, which has no overhead comparing to similar approaches like SMILES \cite{weininger1998JCICS}, which generates a text sequence from the original graph. For the training part, authors were not only interested in generating chemically valid compounds, but also tried to optimize the generation process toward some non-differentiable metrics (e.g., how likely the new molecule is water-soluble or fat-soluble) using RL agent. In Molecular GAN (MolGAN), an external software will compute the RL loss for each molecule. The linear combination of RL loss and WGAN loss is utilized by the generator.

Weininger et al. in \cite{weininger1998JCICS} tackled the same problem. Comparing to \cite{de2018arXiv}, they encoded the molecules as text sequences by using SMILES, the string representation of the molecule, not the original graph-based one. They presented Objective-Reinforced Generative Adversarial Networks (ORGAN), which is built on SeqGAN \cite{yu2017AAAI} and their RL agent uses REINFORCE \cite{williams1992ML}, a gradient-based approach instead of deep deterministic policy gradient (DDPG) \cite{lillicrap2016ICLR}, an off-policy actor-critic algorithm which Cao et al. used in \cite{de2018arXiv}.
MolGAN gains better chemical property scores comparing to ORGAN, but this model suffers from mode collapse because both the GAN and the RL objective do not encourage generating diverse outputs; alternatively, the ORGAN RL agent depends on REINFORCE, and the unique score is optimized penalizing non-unique outputs.

\begin{figure*}[t]
  \centering
    \begin{subfigure}[t]{\columnwidth}
           \centering
            \includegraphics[scale=0.3]{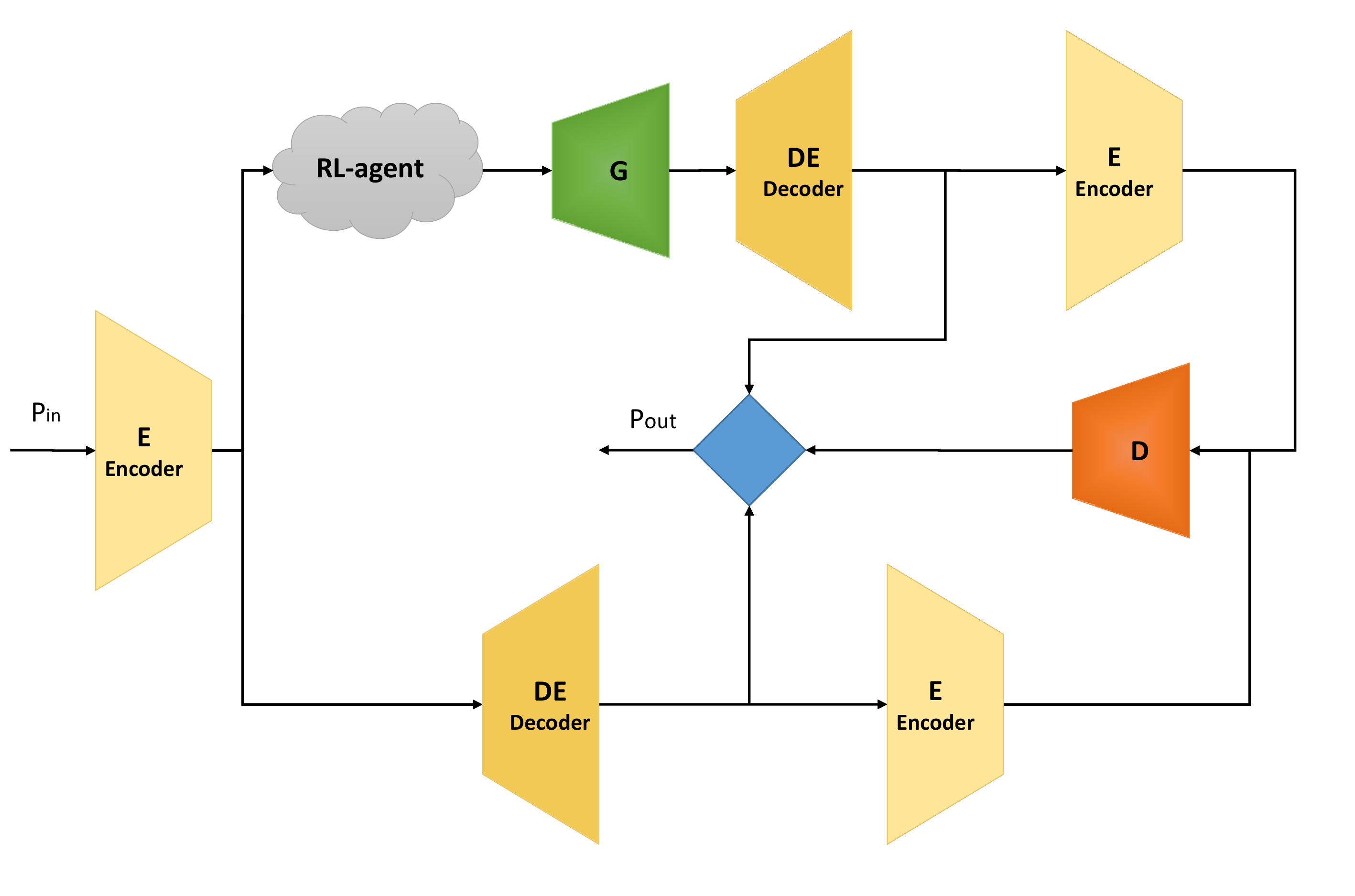}
           \caption{RL-GAN-Net\cite{sarmad2019CVPR}}
    \end{subfigure}
    \begin{subfigure}[t]{\columnwidth}
           \centering
           \includegraphics[scale=0.35]{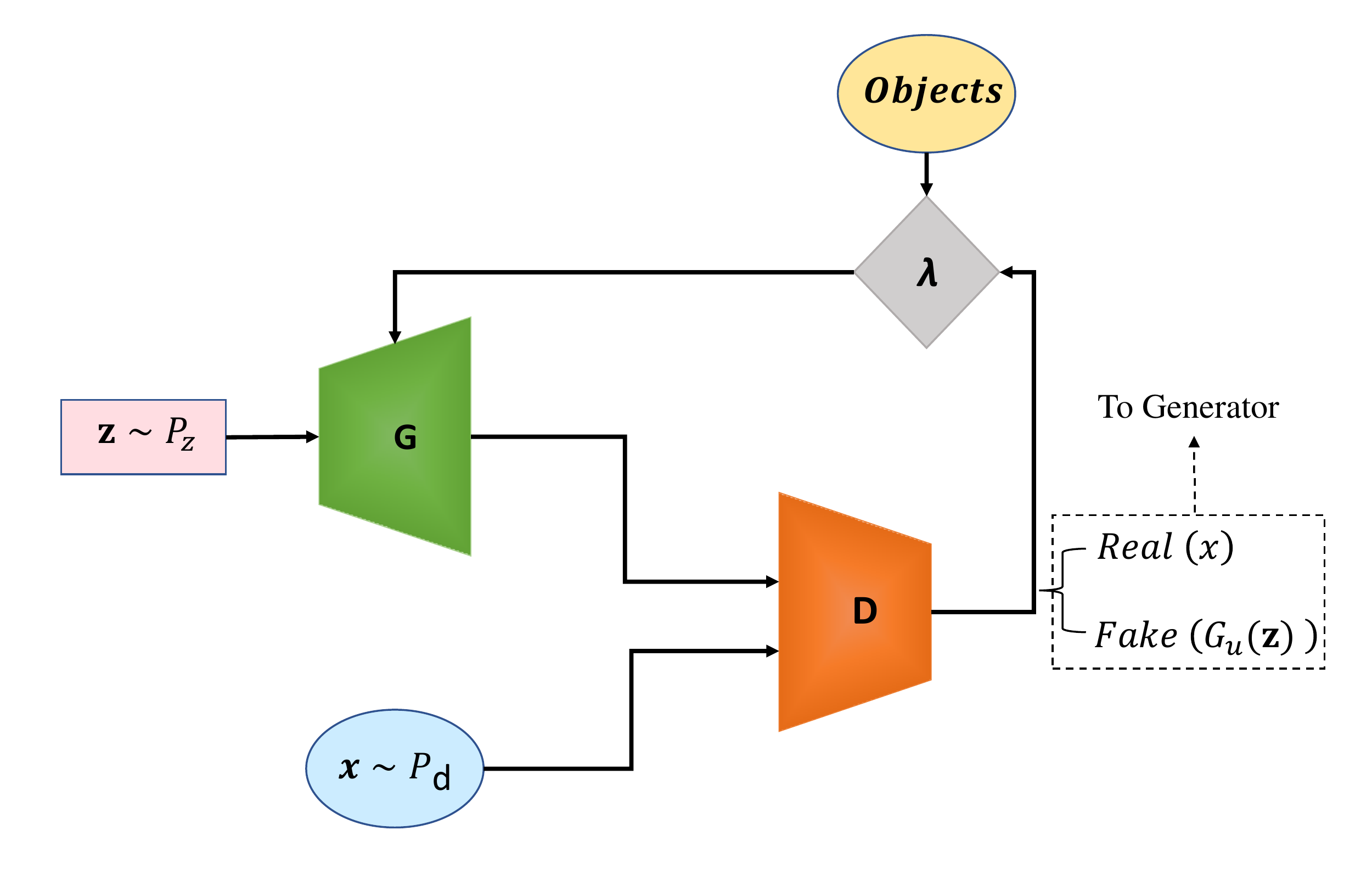}
           \caption{ORGAN \cite{guimaraes2017arXiv}}
    \end{subfigure}
    \begin{subfigure}[t]{\columnwidth}
          \centering
            \includegraphics[scale=0.35]{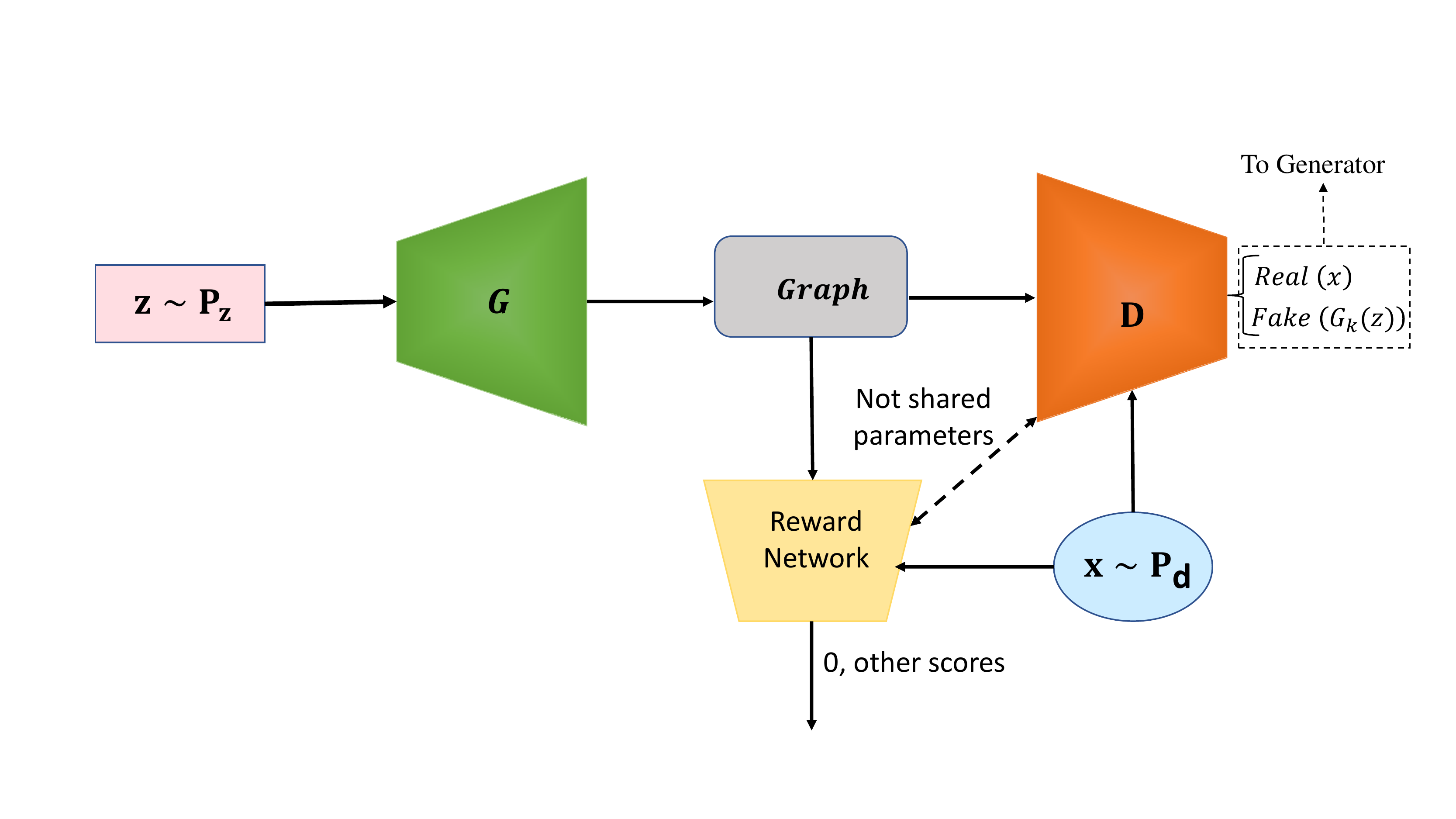}
           \caption{MolGAN \cite{de2018arXiv}}
    \end{subfigure}
     \caption{Schematic view of GANs with one generator, one discriminators, and one RL agent reviewed in Subsection~\ref{SubSec:One_generator_One_discriminator_One_RL_agent}. 
    In (b) G is trained to maximize a weighted average of the objective, and the discriminator.
     In (c) the reward network approximate the reward function of a sample and optimize molecule generation using RL. It assigns a reward to each molecule to match a score provided by an external software.
 $D$: Discriminator, $G$: Generator, $z$: Noise, $x$: Real data, $P_z$: Latent space, $P_x$: Data distribution, $G(z)$: Fake data generated by the generator, $Objectives$: domain-specific metrics}
     \label{Fig:Architecures_Sec4.2.5}
\end{figure*}

For controlling the generator, we can also use an RL agent. In \cite{sarmad2019CVPR}, Sarmad et al.  presented RL-GAN-Net, a real-time completion framework for point cloud shapes. Their suggested architecture is the combination of an auto-encoder (AE), a reinforcement learning (RL) agent and a latent-space generative adversarial network (l-GAN). Based on the pre-trained AE, the RL agent selects the proper seed for the generator. This idea of controlling the GAN's output can open up new potentialities to overcome the fundamental instabilities of current deep architectures. Fig.~\ref{Fig:Architecures_Sec4.2.5} presents the structure of GAN variants, described in this section. 
\begin{table*}[ht!]
  \centering
  \caption{Summary of the publication included in the modified architecture category (Subsection \ref{Modified_Architecture})}
  \setlength{\tabcolsep}{4pt}
  \begin{tabularx}{\textwidth}{|p{1.62cm}|p{6.51cm}|p{4.55cm}|X|}
  \hline
	  \rowcolor{LightBlue}
	  Reference & Methodology and Contributions & Pros & Cons \\ \hline 
	 \rowcolor{LightGray}
	 \multicolumn{4}{|c|}{Multiple generators, One discriminator: Subsection \ref{SubSec:Multiple_generators_One_discriminator} } \\ \hline
	Stackelberg GAN\cite{zhang2018arXiv} & Tackling the instability problem  in the training procedure with multi-generator architecture
    &More stable training performances, alleviate the mode collapse &-\\ \hline
	MADGAN\cite{ke2020IEEE} &A multiagent distributed GAN framework based on the social group wisdom
     &Simultaneously training of multiple generators, consistency of all generators’ training results
     &-   \\ \hline
	MAD-GAN \cite{ghosh2018IEEE}  & A multi-agent diverse GAN architecture & Capturing diverse modes while producing high-quality samples & Assumption of infinite capacity for players, global optimal is not practically reachable   \\ \hline
	MGAN \cite{hoang2018ICLR} & Encouraging generators to generate separable data by classifier & Overcoming the mode collapsing, diversity & -  \\ \hline
	MPMGAN\cite{ghosh2016arXiv} & An innovative message passing model, where messages being passed among generators & Improvement in image generation, valuable representations from message generator &- \\ \hline
	 \rowcolor{LightGray}
    \multicolumn{4}{|c|}{One generator, Multiple discriminators: Subsection \ref{SubSec:One_generator_Multiple_discriminators} } \\ \hline
     DDL-GAN \cite{jin2020ICME} &  Using DDL & Diversity & Only applicable to multiple discriminators  \\ \hline
     D2GAN \cite{nguyen2017ANIPS}  & Combine KL, reverse KL  & Quality and diversity, scalable for large-scale datasets & Not powerfull as the combination of autoencoder or GAN  \\ \hline
     GMAN \cite{durugkar2016arXiv} & Multiple discriminators & Robust to mode collapse & Complexity, converge to same outputs \\ \hline
     microbatch GAN \cite{mordido2020IEEE}  & Using microbatch & Mitigate mode collapse & -  \\ \hline
     MD-GAN \cite{hardy2019IPDPS} & Parallel computation, distributed data & Less communication cost, computation complexity & -  \\ \hline
     FakeGAN \cite{aghakhani2018SPW}  & Text classification & - & -  \\ \hline
	 \rowcolor{LightGray}
     \multicolumn{4}{|c|}{Multiple generators, Multiple discriminators: Subsection \ref{SubSec:Multiple_generators_Multiple_discriminators} } \\ \hline
     \cite{arora2017arXiv} &Tackling generalization and equilibrium in GANs
     & Improve the quality of several existing GAN training methods & Aggregation of losses with an extra regularization term, discourages the weights being too far away from uniform \\ \hline
     MADGAN\cite{ke2020IEEE} & Address the multiagent cognitive consistency problem in large-scale distributed network &Simultaneously training of multiple generators, consistency of all generators’ training results &- \\ \hline 
     FedGAN\cite{rasouli2020arXiv}& A multi-generator and multi-discriminator architecture for training a GAN with distributed sources &Similar performance to general distributed GAN with reduction in communication complexity &- \\ \hline
	 \rowcolor{LightGray}
	 \multicolumn{4}{|c|}{One generators, One discriminator, One classifier: Subsection \ref{SubSec:One_generator_One_discriminator_One_classifier} } \\ \hline
	 CS-GAN \cite{li2018Elsevier}
	 & Combile RNN, GAN, and RL, Use a classifier to validate category, A character level model
	 & generate sentences based on category, limiting action space
	 & - \\ \hline
 	 \cite{tran2019ANIPS}
	 & Multi-class minimax game based self-supervised tasks,
	 & Improve convergence, can integrate into GAN models
	 & -\\ \hline
	 \rowcolor{LightGray}
	 \multicolumn{4}{|c|}{One generators, One discriminator, One RL agent: Subsection \ref{SubSec:One_generator_One_discriminator_One_RL_agent} } \\ \hline
	 MolGAN \cite{de2018arXiv}
	 & Use Original graph-structured data,use RL objective to generate specific chemical property
	 & Better chemical property scores, no overhead in representation
	 & Susceptible to mode collapse \\ \hline
	 ORGAN \cite{guimaraes2017arXiv}
	 & Encode molecules as text sequences, control properties of generated samples with RL, use Wasserstein distance as loss function
	 & Better result than trained RNNs via MLE or SeqGAN
	 & Overhead in representation, works only on sequential data \\ \hline
	 RL-GAN-Net \cite{sarmad2019CVPR}
	 & Use RL to find correct input for GAN, Combine AE, RL and l-GAN
	 & A real time point cloud shape completion, less complexity
	 & - \\ \hline
  \end{tabularx}
  \label{Tab:Summery_Modified_Architecture}
\end{table*}
\subsection{Modified Learning Algorithm}
\label{Subsec:Modified_Learning_Algorithm}
This category covers methods in which the proposed improvements involve modification in learning methods. Here, in this section, we turn our attention to the literatures which combine the other learning approaches such as fictitious play and reinforcement learning with GANs.

Different variation of GANs which are surveyed in Section~\ref{SubSec:No-regret_learning} study GAN training process as a regret minimization problem instead of the popular view which seeks to minimize the divergence between real and generated distributions. 
 As another learning method, Section \ref{SubSec:Fictitious_play} utilizes fictitious play to simulate the training algorithm on GAN. Section~\ref{SubSec:Federated_learning} provides a review on the proposed GAN models that are used a federated learning framework which trains across distributed sources  to overcome the data limitation of GANs.
 Researches in Section~\ref{SubSec:RL} seek to make a connection between GAN and RL. Table \ref{Tab:Summery_Modified_Learning} summarizes the contributions, pros and limitations of literature reviewed in this section.

\subsubsection{No-regret learning}
\label{SubSec:No-regret_learning}

The best response algorithms for GAN are often computationally intractable, and they do not lead to convergence and have cycling behavior even in simple games. However, the simple solution, in that case, is to average the iterates. Regret minimization is the more suitable way to think about GAN training dynamics. In \cite{kodali2017arXiv}, Kodali et al. proposed studying GAN training dynamics as a repeated game that both players use no-regret algorithms. Moreover, the authors showed that the GAN game’s convex-concave case has a unique solution. If G and D have enough capacity in the non-parametric limit and updates are made in the function space, the GAN game is convex-concave. It also can be guaranteed convergence (of averaged iterates) using no-regret algorithms. With standard arguments from game theory literature, the authors show that the discriminator does not need to be optimal at each step.

In contrast to \cite{kodali2017arXiv}, much of the recent developments \cite{goodfellow2016arXiv} are based on the unrealistic assumption that the discriminator is playing optimally; this corresponds to at least one player using the best-response algorithm. But in the practical case with neural networks, these convergence results do not hold because the game objective function is non-convex. In non-convex games, global regret minimization and equilibrium computation are computationally hard.
Moreover, Kodali et al. in \cite{kodali2017arXiv} also analyzed GAN training's convergence from this point of view to understand mode collapse. They showed that mode collapse happens because of undesirable local equilibria in this non-convex game (accompanied by sharp gradients of the discriminator function around some real data points). Furthermore, the authors showed that a gradient penalty scheme can avoid the mode collapse by regularizing the discriminator to constrain its gradients in the ambient data space.

In \cite{grnarova2017arXiv}, Grnarova et al. used regret minimization. They provided a method that provably converges to an MN equilibrium. Because the minimax value of pure strategy for the generators is always higher than the minimax value of the mix equilibrium strategy of generators; thus, the generators are more suitable. This convergence happens for semi-shallow GAN architectures using regret minimization procedures for every player. Semi-shallow GAN architectures are architectures that the generator is any arbitrary network, and the discriminator consists of a single layer network.
This method is done even though the game induced by such architectures is not convex-concave. Furthermore, they show that the minimax objective of the generator's equilibrium strategy is optimal.


\subsubsection{Fictitious play}
\label{SubSec:Fictitious_play}

GAN is a two-player zero-sum game with a repeated game as the training process. If the zero-sum game is played repeatedly between two rational players, they try to increase their payoff. Let $s_i^n \in S_i$ show the action taken by player $i$ at time $n$ and $\{s_i^0, s_i^1, ... , s_i^{n-1}\}$ are previous actions chosen by player $i$. So player $j$ can choose the best response, assuming player $i$ is choosing its strategy according to the empirical distribution of $\{s_i^0, S_i^1, ... , s_i^{n-1}\}$. Thus, the expected utility is a linear combination of utilities under different pure strategies. So we can assume that each player plays the best pure response at each round.
In game theory, this learning rule is called fictitious play and can help us find the Nash equilibrium. 
The fictitious play achieves a Nash equilibrium in two-player zero-sum games if the game's equilibrium is unique. However, if there exist multiple Nash equilibria, other initialization may yield other solutions.

By relating GAN with the two-player zero-sum game, Ge et al. designed a training algorithm to simulate the fictitious play on GAN and provide a theoretical convergence guarantee in \cite{ge2018ECCV}. They also showed that by assuming the best response at each update in Fictitious GAN, the distribution of the mixture outputs from the generators converges to the data distribution. The discriminator outputs converge to the optimum discriminator function. The authors in \cite{ge2018ECCV} used two queues D and G, to store the historically trained models of the discriminator and the generator. They also showed that Fictitious GAN can effectively resolve some convergence issues that the standard training approach cannot resolve and can be applied on top of existing GAN variants.

\subsubsection{Federated learning}
\label{SubSec:Federated_learning}

Data limitation is a common drawback in deep learning models like GANs. We can solve this issue by using distributed data from multiple sources, but this is difficult due to some reasons like privacy concerns of users, communication efficiency and statistical heterogeneity, etc. This brings the idea of using federated learning in GANs to address these subjects \cite{rasouli2020arXiv, fan2020arXiv}.

Rasouli et al. in \cite{rasouli2020arXiv}, proposed a federated approach to GANs, which trains over distributed sources with non-independent-and-identically-distributed data sources. In this model, every K time steps of local gradient, agents send their local discriminator and generator parameters to the intermediary and receive back the synchronized parameters. Due to the average communication per round per agent, FedGAN is more efficient compare to general distributed GAN. Experiments also proved FedGAN is robust by increasing K.
For proving the convergence of this model, the authors connect the convergence of GAN to convergence of an Ordinary Differential Equation (ODE) representation of the parameter updates \cite{mescheder2017ANIPS} under equal or two time-scale updates of generators and discriminators.
Rasouli et al. showed that the FedGAN ODE representation of parameters update asymptotically follows the ODE representing the parameter update of the centralized GAN. So by using the existing results for centralized GAN, FedGAN also converges.

Fan et al. also proposed a generative learning model using a federated learning framework in \cite{fan2020arXiv}. The aim is to train a unified central GAN model with the combined generative models of each client.
Fan et al. examined 4 kinds of synchronization strategies, synchronizing each the central model of D and G to every client (Sync D\&G) or simply sync the generator or the discriminator (Sync G or Sync D) or none of them (Sync None). In situations where communication costs are high, they recommend Sync G while losing some generative potential, otherwise synchronize both D and G.
\cite{fan2020arXiv} results showed that federate learning is commonly robust to the number of agents with Independent and Identically Distributed (IID) and fairly non-IID training data. However, for highly skewed data distribution, their model performed abnormality due to weight divergence.


\subsubsection{Reinforcement learning}
\label{SubSec:RL}

Cross-modal hashing tries to map different multimedia data into a common Hamming space, realizing fast and flexible retrieval across different modalities. Cross-modal hashing has two weaknesses: (1) Depends on large-scale labeled cross-modal training data. (2) Ignore the rich information contained in a large amount of unlabeled data across different modalities. So Zhang et al. in \cite{zhang2018cybernetics} proposed Semi-supervised Cross-modal Hashing GAN (SCH-GAN) that exploits a large amount of unlabeled data to improve hashing learning. The generator takes the correlation score predicted by the discriminator as a reward and tries to pick margin examples of one modality from unlabeled data when giving another modality query. The discriminator tries to predict the correlation between query and chosen examples of the generator using Reinforcement learning.

An agent trained using RL is only able to achieve the single task specified via its reward function. So Florensa et al. in \cite{florensa2018PMLR} provided Goal Generative Adversarial Network (Goal GAN). This method allows an agent to automatically discover the range of tasks at the appropriate level of complexity for the agent in its environment with no prior knowledge about the environment or the tasks being performed and allows an agent to generate its own reward functions. The goal discriminator is trained to evaluate whether a goal is at the appropriate level of difficulty for the current policy. The goal generator is prepared to generate goals that meet these criteria.

GAN has limitations when the goal is for generating sequences of discrete tokens. First, it is hard to provide the gradient update from the discriminator to the generator when the outputs are discrete. Second, The discriminator can only reward an entire sequence after generation; for a partially generated sequence, it is non-trivial to balance how well it is now and how well it will be in the future as the whole sequence. Yu et al. in \cite{yu2017AAAI} proposed Sequence GAN (SeqGAN) and model the data generator as a stochastic policy in reinforcement learning (RL). The RL reward signal comes from the discriminator decided on a complete sequence and, using the Monte Carlo search, is passed back to the intermediate state-action steps. So in this method, they care about the long-term reward at every timestep. The authors consider not only the fitness of previous tokens but also the resulted future outcome. "This is similar to playing the games such as Go or Chess, where players sometimes give up the immediate interests for the long-term victory" \cite{silver2016}.

The main problem in \cite{yu2017AAAI} is that the classifier's reward cannot accurately reflect the novelty of text. 
So, in \cite{xu2018EMNLP} in comparison to \cite{yu2017AAAI}, Yu et al. assigned a low reward for repeatedly generated text and high reward for "novel" and fluent text, encouraging the generator to produce diverse and informative text, and propose a novel language-model based discriminator, which can better distinguish novel text from repeated text without the saturation problem. The generator reward consists of two parts, the reward at the sentence level and that at the word level. The authors maximized the reward of real text and minimize fake text rewards to train the discriminator. The reason for minimizing the reward of generated text is that the text that is repeatedly generated by the generator can be identified by the discriminator and get a lower reward. The motivation of maximizing the reward of real-world data lies in that not only the uncommon text in the generated data can get a high reward, but also the discriminator can punish low-quality text to some extent.

The same notion of SeqGAN can be applied in domains such as image captioning. Image captioning's aim is to describe an image with words. Former approaches for image captioning like maximum likelihood method suffer from a so-called exposure bias problem which happens when the model tries to produce a sequence of tokens based on previous tokens. In this situation, the model may generate tokens that were never seen in training data \cite{bengio2015scheduled}. Yan et al. in \cite{yan2018ICPR} used the idea of SeqGAN to address the problem of exposure bias. In this scheme, the image captioning generator is considered as the generator in the GAN framework whose aim is to describe the images. The discriminator has two duties, the first is to distinguish the real description and generated one and the second is to figure out if the description is related to the image or not. To deal with the discreteness of the generated text, the discriminator is considered as an agent which produces a reward for the generator. Although, lack of intermediate reward is another problem which solves by using the Monte Carlo roll-out strategy same as SeqGAN.

Finding new chemical compounds and generating molecules are also challenging tasks in a discrete setting. \cite{guimaraes2017arXiv} and \cite{de2018arXiv} tackled this problem and proposed two models that rely on SeqGAN. The main difference is adding an RL component to the basic architecture of GAN, where we discussed in Section \ref{SubSec:One_generator_One_discriminator_One_RL_agent}.

The idea behind SeqGAN has also been applied to generating sentences with certain labels. Li et al. in \cite{li2018Elsevier} introduced CS-GAN, which consists of a generator and a descriptor (discriminator and classifier). In this model, the generator takes an action, and the descriptor task is to identify sentence categories by returning the reward. Details of this model are explained in Section~\ref{SubSec:One_generator_One_discriminator_One_classifier}.

Aghakhani et al. in \cite{aghakhani2018SPW} introduced a system that for the first time expands GANs for a text classification task, specifically, detecting deceptive reviews (FakeGAN).
Previous models for text classification have limitations: (1) Biased problems like Recurrent NN, where later words in a text have more weight than earlier words. (2) Correlation with the window size like CNN.
Unlike standard GAN with a single Generator and Discriminator, FakeGAN uses two discriminators and one generator. The authors modeled the generator as a stochastic policy agent in reinforcement learning (RL) and used the Monte Carlo search algorithm for the discriminators to estimate and pass the intermediate action-value as the RL reward to the generator.  One of the discriminators tries to distinguish between truthful and deceptive reviews, whereas the other tries to distinguish between fake and real reviews.

Ghosh et al. in \cite{ghosh2016arXivHandwriting} used GANs for learning the handwriting of an entity and combine it with reinforcement learning techniques to achieve faster learning. The generator can generate words looking similar to the reference word, and the discriminator network can be used as an OCR (optical character recognition) system.
The concept of reinforcement learning comes into play when letters need to be joined to form words, such as the spacing between characters and strokes from one note to another, and provide suitable rewards or penalties for the generator to learn the handwriting with greater accuracy.

The optimized generation of sequences with particular desired goals is challenging in sequence generation tasks. Most of the current work mainly learns to generate outputs that are close to the real distribution. However, in many applications, we need to generate data similar to real ones and have specific properties or attributes. 
Hossam et al. in \cite{hossam2020arXiv} introduced the first GAN-controlled generative model for sequences that address the diversity issue in a principled approach. The authors combine GAN and RL policy learning benefits while avoiding mode-collapse and high variance drawbacks. 
The authors show that if only pure RL is applied with the GAN-based objective, the realistic quality of the output might be sacrificed for the cause of achieving a higher reward. For example, in the text-generation case, by generating sentences in which few words are repeated all the time, the model could achieve a similar quality score. Hence, combining a GAN-based objective with RL promotes the optimization process of RL to stay close to the actual data distribution. 
This model can be used for any GAN model to enable it to optimize the desired goal according to the given task directly.

A novel RL-based neural architecture search (NAS) methodology is proposed for GANs in \cite{tian2020Springer} by Tian et al.
Markov decision process formulation is applied to redefine the issue of neural architecture search for GANs in this article, therefore a more effective RL-based search algorithm with more global optimization is achieved.
Additionally, data efficiency can be improved due to better facilitation of off-policy RL training by this formulation \cite{tian2020Springer}.
On-policy RL is used in most of the formerly proposed search methods employed in RL-based GAN architecture, which may have a significantly long training time because of limited data efficiency. Agents in off-policy RL algorithms are enabled to learn more accurately as these algorithms use past experience. However, using off-policy data can lead to unstable policy network training because these training samples are systematically different from the on-policy ones \cite{tian2020Springer}. A new formulation in \cite{tian2020Springer} supports the off-policy strategy better and lessens the instability problem.



It is worth noting that authors in the reviewed papers have used different metrics to evaluate their proposed GAN models. Table~\ref{Tab:Summery_Metrics2} presents all these metrics and indicates which ones were used in each of the papers. Metrics range from quantitative to qualitative ones. The most-reported metrics are Inception Scores and FID. 
\begin{table*}[ht!]
  \centering
  \caption{Summary of the publication included in the learning method category (Subsection \ref{Modified_Architecture}).}
  \setlength{\tabcolsep}{3pt}
  \begin{tabularx}{\textwidth}{|p{1.65cm}|p{5.5cm}|p{5.5cm}|X|}
  \hline
	  \rowcolor{LightBlue}
	  Reference & Methodology and Contributions & Pros & Cons \\ \hline 
	 \rowcolor{LightGray}
	 \multicolumn{4}{|c|}{No-regret learning: Subsection \ref{SubSec:No-regret_learning} } \\ \hline
	 
     DRAGAN \cite{kodali2017arXiv} & Applying no-regret algorithm, new regularizer & High stability across objective functions, mitigates mode collapse & - \\ \hline
     Chekhov GAN \cite{grnarova2017arXiv} & Online learning algorithm for semi concave game & Converge to mixed NE for semi shallow discriminator & - \\ \hline
	 \rowcolor{LightGray}
	 \multicolumn{4}{|c|}{Fictitious play: Subsection \ref{SubSec:Fictitious_play} } \\ \hline
	 
	 Fictitious GAN \cite{ge2018ECCV} & Fictitious (historical models) &
     Solve the oscillation behavior, solve divergence issue on some cases, applicable & Applied only on 2 player zero-sum games \\ \hline
	 \rowcolor{LightGray}
     \multicolumn{4}{|c|}{Federated learning: Subsection \ref{SubSec:Federated_learning} } \\ \hline
     FedGAN \cite{rasouli2020arXiv} 
     & Communication-efficient distributed GAN subject to privacy constraints, connect the convergence of GAN to ODE 
     & Prove the convergence, less communication complexity compare to general distributed GAN
     & - \\ \hline
     \cite{fan2020arXiv}
     & Using a federated learning framework
     & Robustness to the number of clients with IID and moderately non-IID data
     & Performs anomaly for highly skewed data distribution, accuracy drops with non-IID data \\ \hline
	 \rowcolor{LightGray}
	 \multicolumn{4}{|c|}{Reinforcement learning: Subsection \ref{SubSec:RL} } \\ \hline
	 \cite{florensa2018PMLR}& Generate diverse appropriate level of difficulty set of goal & - & - \\ 	 \hline
	 Diversity-promoting GAN \cite{xu2018EMNLP}& New objective function, generate text & Diversity and novelty & - \\ 	 \hline
	 \cite{zhang2018IEEE}& Using GAN for cross-model hashing & Extract rich information from unlabeled data & - \\ 	 \hline
	 SeqGAN \cite{yu2017AAAI}& Extending GANs to generate sequence of discrete tokens & Solve the problem of discrete data & - \\
	 \hline
	 FakeGAN \cite{aghakhani2018SPW}& Text classification & - & - \\  	 \hline
	 CS-GAN \cite{li2018Elsevier}& Combine RL, GAN, RNN & More realistic, faster & - \\  \hline
	 \cite{ghosh2016arXivHandwriting}& Handwriting recognition & - & - \\ \hline
	 ORGAN  \cite{guimaraes2017arXiv}
	 & RL agent + SeqGAN
	 & Better result than RNN trained via MLE or SeqGAN
	 & Works only on sequential data\\  \hline
	 MolGAN \cite{de2018arXiv}
	 & RL agent + SeqGAN
	 & optimizing non-differentiable metrics by RL \& Faster training time
	 & Susceptible to mode collapse \\  \hline
	 OptiGAN \cite{hossam2020arXiv}& Combining MLE and GAN & Used for different models and goals & -\\ \hline
	 \cite{tian2020Springer}& Redefining the issue of neural architecture search for GANs by applying Markov decision process formulation & More effective RL-based search algorithm, smoother architecture sampling &- \\ \hline
  \end{tabularx}
  \label{Tab:Summery_Modified_Learning}
\end{table*}
\begin{table*}[t]
  \centering
  \caption{Metrics which are reported for results in the reviewed papers}
  \setlength{\tabcolsep}{4pt}
  \begin{tabularx}{\textwidth}{|p{4cm}|X|} 
  \hline
      \rowcolor{LightBlue}
	  Metrics & Reference  \\  \hline
	  Inception Score & Stackelberg GAN \cite{zhang2018arXiv}, \cite{farnia202arXiv},       \cite{franci2020arXiv}, MGAN \cite{hoang2018ICLR},DDL-GAN \cite{jin2020ICME}, GMAN \cite{durugkar2016arXiv}, MD-GAN \cite{hardy2019IPDPS}, \cite{arora2017arXiv}, DRAGAN \cite{kodali2017arXiv}, Fictitious GAN \cite{ge2018ECCV},  \cite{tian2020Springer}, D2GAN \cite{nguyen2017ANIPS}\\ \hline
	  FID & Stackelberg GAN \cite{zhang2018arXiv}, DDL-GAN \cite{jin2020ICME}, Microbatch GAN \cite{mordido2020IEEE}, MD-GAN \cite{hardy2019IPDPS},  \cite{tran2019ANIPS}, FedGAN \cite{rasouli2020arXiv},   \cite{tian2020Springer}\\ \hline
	  Chi-square &  MAD-GAN \cite{ghosh2018IEEE}  \\ \hline
	  KL &  MAD-GAN \cite{ghosh2018IEEE}  \\ \hline
	  Reverse KL & Chekhov GAN \cite{grnarova2017arXiv} \\ \hline
	  P-value & SeqGAN \cite{yu2017AAAI} \\ \hline
	  Human Scores & Diversity-promoting GAN \cite{xu2018EMNLP}, SeqGAN \cite{yu2017AAAI}\\ \hline
	  Classification Scores & \cite{fan2020arXiv},  \cite{zhang2018IEEE}, \cite{aghakhani2018SPW}, CS-GAN \cite{li2018Elsevier} \\ \hline
	  Others & \cite{fan2020arXiv}, ORGAN \cite{guimaraes2017arXiv}, OptiGAN \cite{hossam2020arXiv}\\ \hline
	  	\end{tabularx}
\label{Tab:Summery_Metrics2}
\end{table*}

\section{Open Problems and Future Work}
\label{Sec:Discussion}

We conducted this review of recent progresses in GANs using game theory which can serve as a reference for future research. Comparing this survey to the other reviews in the literature, and considering the many published works which deal with GAN challenges, we emphasize on the theory aspects of GAN and not GAN applications. This is all done through the taking of a game theory perspective based on our proposed taxonomy.


Although there are various studies that have explored different aspects of GANs, but, several challenges still remain should be investigated. In this section, we discuss such challenges, especially in the discussed subject, game of GANs, and propose future research directions to tackle these problems.


While GANs achieve the state-of-the-art performance and compelling results on various generative tasks, but, these results come at some challenges, especially difficulty in the training of GANs. Training procedure suffers from instability problems.
While reaching to Nash equilibrium, generator and discriminator are trying to minimize their own cost function, regardless of the other one. This can cause the problem of non-convergence and instability because of minimizing one cost can lead to maximizing the other one's cost.
Another main problem of GANs which needs to be addressed is mode collapse. This problem becomes more critical for unbalanced data sets or when the number of classes is high. In other hand, when discriminator works properly in distinguishing samples, generators gradients vanishes. This problem which is called vanishing gradient should also be considered.
Compared with other generative models, the evaluation of GANs is more difficult.
This is partially due to the lack of appropriate metrics. Most evaluation metrics are qualitative rather than being quantitative. Qualitative metrics such as human examination of samples, is an arduous task and depends on the subject. 
 
 More specifically, as the authors in \cite{cao2018IEEE} expressed one of the most important future direction is to improve theoretical aspects of GANs to solve problems such as model collapse, non-convergence, and training difficulties.
Although there have many works on the theory aspects, most of the current training strategies are based on the optimization theory, whose scope is restricted to local convergence due to the non-convexity, and the utilization of game theory techniques is still in its infancy.
 At present, the game theory variant GANs are limited, and much of them are highly restrictive, and are rarely directly applicable. Hence, there is much room for research in game-based GANs which are involving other game models.
 
From the convergence viewpoint, most of the current training methods converge to a local Nash equilibrium, which can be far from an actual and global NE.
 While there is vast literature on the GAN' training, only few researches such as \cite{hsieh2019PMLR} formulation the training procedure from the mixed NE perspective, and investigation for mixed NE of GANs should be examined in more depth.
 On the other hand, existence of an equilibrium does not imply that it can be easily find by a simple algorithm. 
 In particular, training GANs requires finding Nash equilibria in non-convex games, and computing the equilibria in these game is computationally hard. In the future, we are expected to see more solutions tries to make GAN training more stable and converge to actual NE.
 
 Multi-agent models such as \cite{zhang2018arXiv, ghosh2018IEEE, hoang2018ICLR, ghosh2016arXiv, durugkar2016arXiv, jin2020ICME, hardy2019IPDPS, aghakhani2018SPW, mordido2020IEEE, nguyen2017ANIPS, arora2017arXiv, rasouli2020arXiv, ke2020IEEE} are computationally more complex and expensive than two-player models and this factor should be taken into account in development of such variants. Moreover, in multi-generator structures the divergence among the generators should be considered such that all of them do not generate the same samples. 

One of the other directions that we expect to witness the innovation in the future is integrating GANs with other learning methods.
There are a variety of methods in multi-agent learning literature which should be explored as they may be useful when applying in the multi-agent GANs.
In addition, it looks like much more research on the relationship and combination between GANs and current applied learning approach such as RL is still required, and it also will be a promising research direction in the next few years.
Moreover, GAN is proposed as unsupervised learning, but adding a certain number of labels, specially in practical applications, can substantially improve its generating capability. Therefore, how to combine GAN and semi-supervised learning is also one of the potential future research topics.

As the final note, GAN is a relatively novel and new model with significant recent progress, so the landscape of possible applications remains open for exploration.
The advancements in solving the above challenges can be decisive for GANs to be more employed in real scenarios.

\balance
\bibliographystyle{IEEEtran}
\bibliography{IEEEabrv,MyReferences}

\end{document}